\documentclass[runningheads]{llncs}
\usepackage{graphicx}
\usepackage{comment}
\usepackage{amsmath,amssymb} % define this before the line numbering.
\usepackage{caption}
\usepackage{color}

\usepackage[width=122mm,left=12mm,paperwidth=146mm,height=193mm,top=12mm,paperheight=217mm]{geometry}

\usepackage[pagebackref=true,breaklinks=true,colorlinks,bookmarks=false]{hyperref}

\definecolor{Yellow}{rgb}{1, 1, 0.6}
\definecolor{Red}{rgb}{1, 0.6, 0.6}
\definecolor{Blue1}{rgb}{0, 0.6, 1}
\definecolor{Red1}{rgb}{0.5, 0.2, 0.2}
\definecolor{PaleYellow}{rgb}{0.8, 0.8, 0}

\newcommand{\jiawen}[1]{}
\newcommand{\tianfan}[1]{}
\newcommand{\xide}[1]{}

\begin{document}
% \renewcommand\thelinenumber{\color[rgb]{0.2,0.5,0.8}\normalfont\sffamily\scriptsize\arabic{linenumber}\color[rgb]{0,0,0}}
% \renewcommand\makeLineNumber {\hss\thelinenumber\ \hspace{6mm} \rlap{\hskip\textwidth\ \hspace{6.5mm}\thelinenumber}}
% \linenumbers
\pagestyle{headings}
\mainmatter
\def\ECCVSubNumber{575}  % Insert your submission number here

\title{Joint Bilateral Learning for Real-time Universal Photorealistic Style Transfer} % Replace with your title

% INITIAL SUBMISSION 
\begin{comment}
\titlerunning{ECCV-20 submission ID \ECCVSubNumber} 
\authorrunning{ECCV-20 submission ID \ECCVSubNumber} 
\author{Anonymous ECCV submission}
\institute{Paper ID \ECCVSubNumber}
\end{comment}
%******************

% CAMERA READY SUBMISSION
%\begin{comment}
\titlerunning{Joint Bilateral Learning for Real-time Universal Photorealistic Style Transfer}
% If the paper title is too long for the running head, you can set
% an abbreviated paper title here
%
\author{Xide Xia\inst{1*}  \and
Meng Zhang\inst{2**} \and
Tianfan Xue\inst{3} \and
Zheng Sun\inst{3} \and
Hui Fang\inst{3**} \and
Brian Kulis\inst{1} \and
Jiawen Chen\inst{3}}
\authorrunning{X. Xia et al.}
% First names are abbreviated in the running head.
% If there are more than two authors, 'et al.' is used.
%
\institute{Boston University \\
\email{\{xidexia,bkulis\}@bu.edu}\\\and
PixelShift.AI\\
\email{meng@pixelshift.ai}\\\and
Google Research\\
\email{\{tianfan,zhengs,hfang,jiawen\}@google.com}}
%\end{comment}
%******************

\maketitle

\makeatletter
\begin{figure}
 \vspace*{-25pt}
    \centering
    \includegraphics[width=\linewidth]{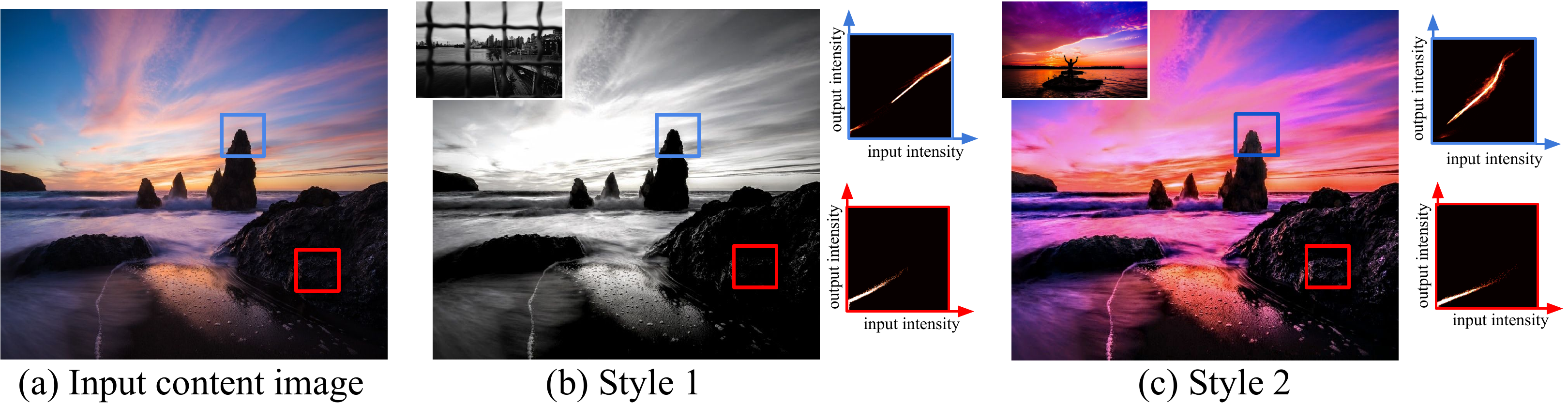}
    \caption{
    Our method takes an input image (a) and renders it in the style of an arbitrary reference photo (insets) while preserving scene content.
    Notice that although the stylized outputs (b) and (c) differ dramatically in appearance from the input, one property they share is that nearby pixels of similar color are transformed similarly.
    We visualize this in the grayscale intensity domain, where the transformation is approximately a curve (zoomed insets).
    Taking advantage of this property, we propose a feed-forward neural network that directly learns these local curves.
	}
    \label{fig:scatter_curve_plot}
     \vspace*{-27pt}
\end{figure}
\makeatother

\begin{abstract}
Photorealistic style transfer is the task of transferring the artistic style of an image onto a content target, producing a result that is plausibly taken with a camera. Recent approaches, based on deep neural networks, produce impressive results but are either too slow to run at practical resolutions, or still contain objectionable artifacts. We propose a new end-to-end model for photorealistic style transfer that is both fast and inherently generates photorealistic results. The core of our approach is a feed-forward neural network that learns local edge-aware affine transforms that automatically obey the photorealism constraint. When trained on a diverse set of images and a variety of styles, our model can robustly apply style transfer to an arbitrary pair of input images. Compared to the state of the art, our method produces visually superior results and is three orders of magnitude faster, enabling real-time performance at 4K on a mobile phone. We validate our method with ablation and user studies.\let\thefootnote\relax\footnotetext{* Work done while interning at Google Research.}\let\thefootnote\relax\footnotetext{** Work done while working at Google Research.}
\end{abstract}

\section{Introduction}
Image style transfer has recently received significant attention in the computer vision and machine learning communities~\cite{jing2019neural}.
A central problem in this domain is the task of transferring the style of an arbitrary image onto a \emph{photorealistic} target.
The seminal work of Gatys et al.~\cite{gatys2016image} formulates this general artistic style transfer problem as an optimization that minimizes both style and content losses, but results often contain spatial distortion artifacts. Luan et al.~\cite{luan2017deep} seek to reduce these artifacts by adding a \emph{photorealism constraint}, which encourages the transformation between input and output to be \emph{locally affine}.
However, because the method formulates the problem as a large optimization whereby the loss over a deep network must be minimized for every new image pair, performance is limited.
The recent work of Yoo et al.~\cite{yoo2019photorealistic} proposes a wavelet corrected transfer based method which provides stable stylization but is not fast enough to run at practical resolutions.
Another line of recent work seeks to pretrain a feed-forward deep model~\cite{dumoulin2017learned,huang2017arbitrary,johnson2016perceptual,li2019learning,li2017diversified,li2018closed,ulyanov2016texture,ulyanov2017improved} that once trained, can produce a stylized result with a single forward pass at test time.
While these ``universal''~\cite{jing2019neural} techniques are significantly faster than those based on optimization, they may not generalize well to unseen images, may produce non-photorealistic results, and are still to slow to run in real time on a mobile device.

In this work, we introduce a fast end-to-end method for photorealistic style transfer.
Our model is a single feed-forward deep neural network that once trained on a suitable dataset, runs in real-time on a mobile phone at full camera resolution (i.e., 12 megapixels or ``4K")---significantly faster than the state of the art.
Our key observation is that we can guarantee photorealistic results by strictly enforcing Luan et. al's photorealism constraint~\cite{luan2017deep}---locally, regions of similar color in the input must map to a similarly colored region in the output while respecting edges.
Therefore, we design an deep learning algorithm in \emph{bilateral space}, where these local affine transforms can be compactly represented.
We contribute:
\begin{enumerate}
    \item A photorealistic style transfer network that learns local affine transforms. Our model is robust and degrades gracefully when confronted with unseen or adversarial inputs.
    \item An inference implementation that runs in real-time at 4K on a mobile phone.
    \item A bilateral-space Laplacian regularizer eliminates spatial grid artifacts.
\end{enumerate}

\begin{figure}[ht]
\hspace{-0.3cm}
\begin{tabular}{c@{\hspace{0.002\linewidth}}c@{\hspace{0.002\linewidth}}c@{\hspace{0.002\linewidth}}c@{\hspace{0.01\linewidth}}c@{\hspace{0.002\linewidth}}c@{\hspace{0.002\linewidth}}c@{\hspace{0.002\linewidth}}c@{\hspace{0.002\linewidth}}c@{\hspace{0.002\linewidth}}c}
  
  \includegraphics[width = .08\linewidth,height=.12\linewidth]{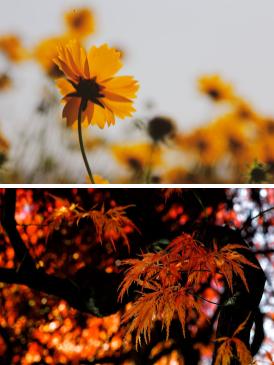} &
  \includegraphics[width = .14\linewidth,height=.12\linewidth]{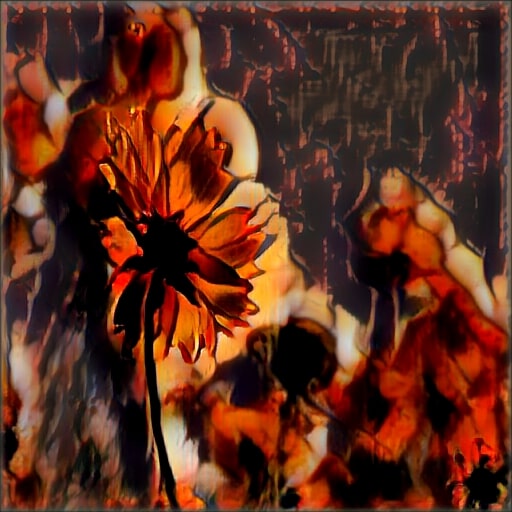} &
  \includegraphics[width = .14\linewidth,height=.12\linewidth]{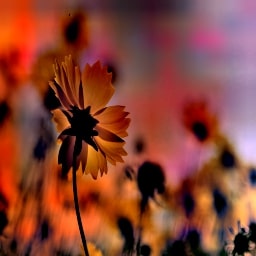} &
  \includegraphics[width = .14\linewidth,height=.12\linewidth]{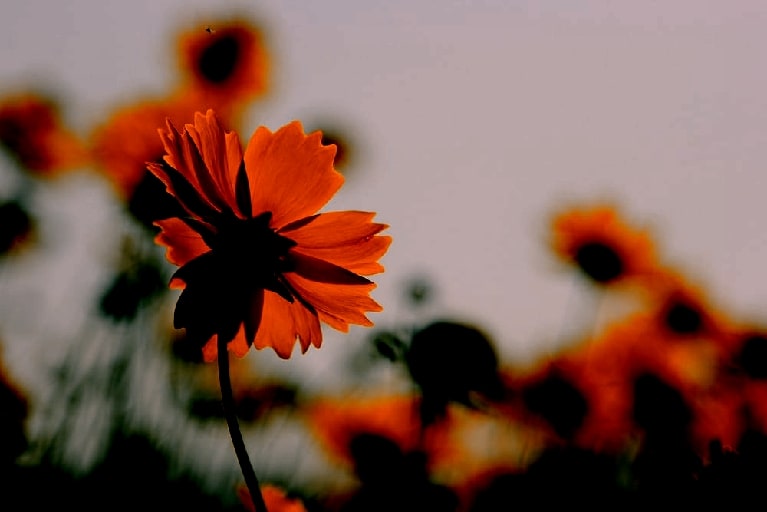} &
  
  \includegraphics[width = .08\linewidth,height=.12\linewidth]{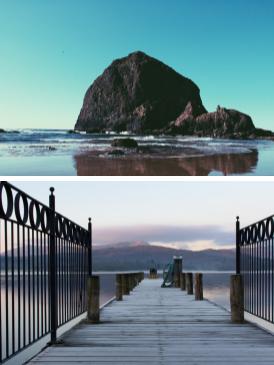} &
  \includegraphics[width = .14\linewidth,height=.12\linewidth]{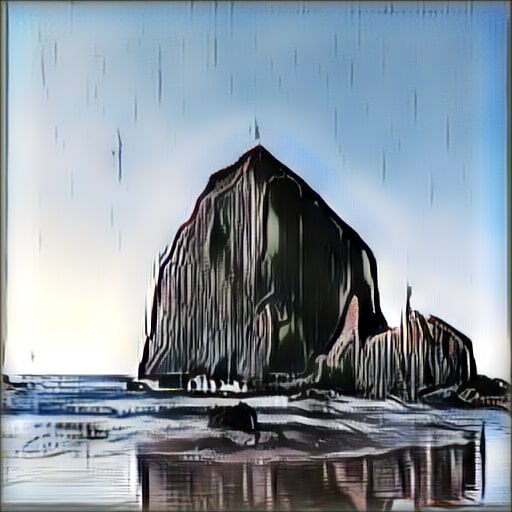} &
  \includegraphics[width = .14\linewidth,height=.12\linewidth]{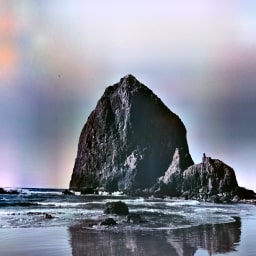} &
  \includegraphics[width = .14\linewidth,height=.12\linewidth]{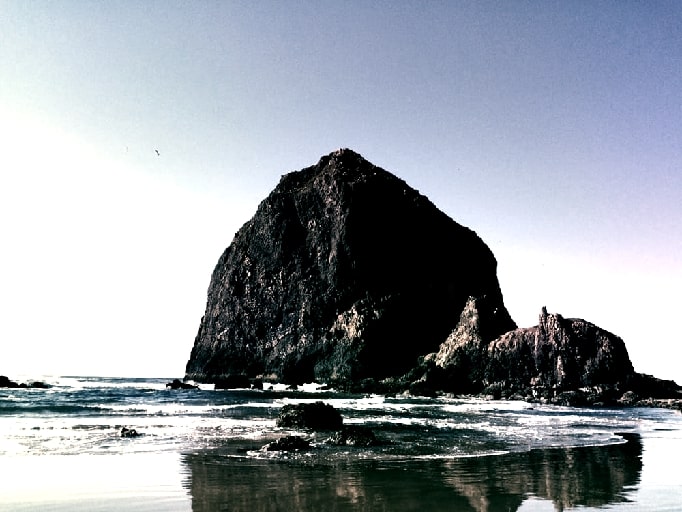} &\\
  
   \includegraphics[width = .08\linewidth,height=.12\linewidth]{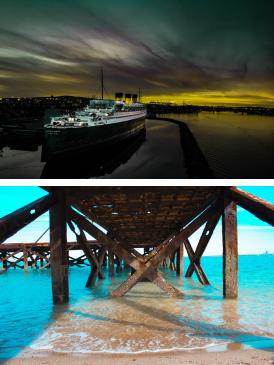} &
  \includegraphics[width = .14\linewidth,height=.12\linewidth]{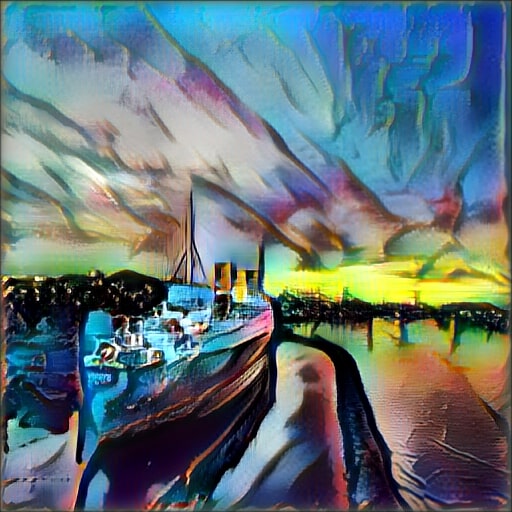} &
  \includegraphics[width = .14\linewidth,height=.12\linewidth]{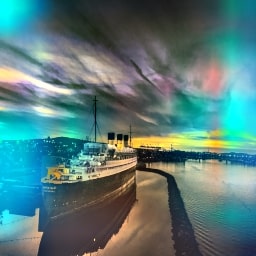} &
  \includegraphics[width = .14\linewidth,height=.12\linewidth]{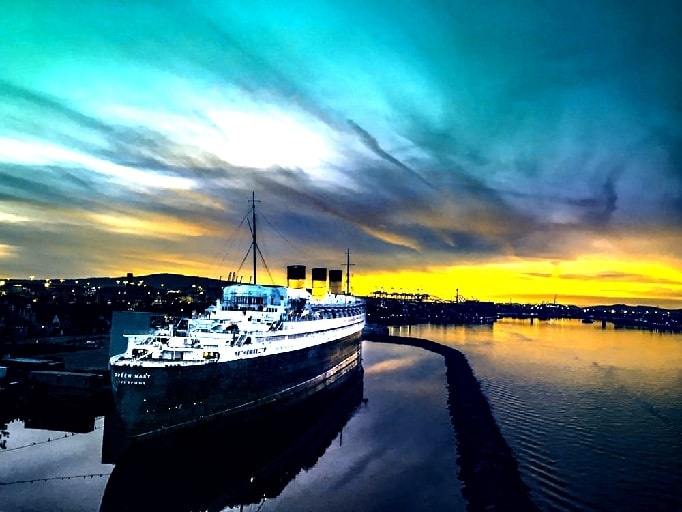} &
  
  \includegraphics[width = .08\linewidth,height=.12\linewidth]{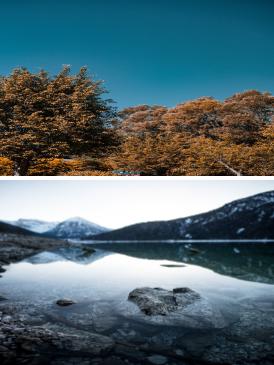} &
  \includegraphics[width = .14\linewidth,height=.12\linewidth]{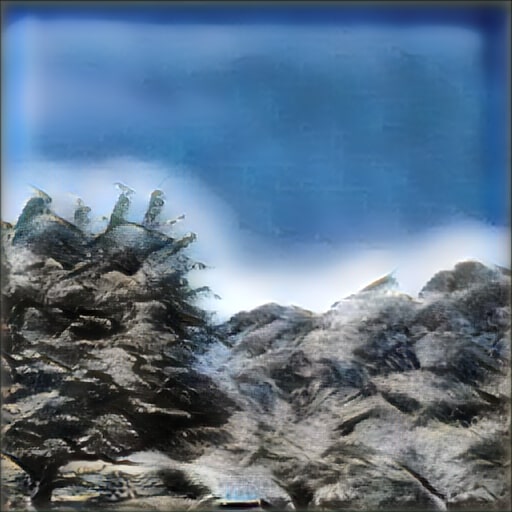} &
  \includegraphics[width = .14\linewidth,height=.12\linewidth]{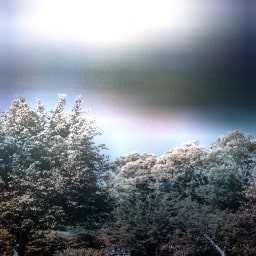} &
  \includegraphics[width = .14\linewidth,height=.12\linewidth]{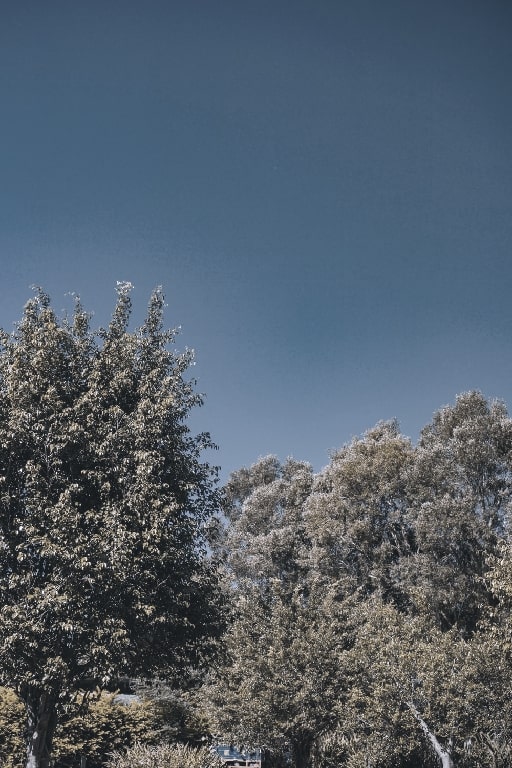} &\\
  
  {Inputs}& {AdaIN}& {HDRnet}& {Ours}& {Inputs}& {AdaIN}& {HDRnet}& {Ours} \\
\end{tabular}
\vspace{-0.3cm}
\caption{\textbf{Inspiration.} Artistic style transfer methods such as AdaIN generalize well to diverse content/style inputs but exhibit distortions on photographic content. HDRnet, designed to reproduce arbitrary imaging operators, learns the transform representation we want but fails to capture universal style transfer. Our work combines ideas from both approaches.}
\label{fig:inspiration}
\vspace{-1.em}
\end{figure}

\subsection{Related Work}

Early work in image style transfer operated by transferring global image statistics~\cite{reinhard2001color} or histograms of filter responses~\cite{pitie2005n}.
As they rely on low-level statistics, they fail to capture semantics.
However, we highlight that these techniques do produce photorealistic results, albeit not always faithful to the style or well exposed.

Recently, Gatys et al.~\cite{gatys2016image} showed that style can be effectively captured by the statistics of layer activations within deep neural networks trained for discriminative image classification.
However, due to its generality, the technique and its successors often contain non-photorealistic painterly spatial distortions.
To remove such distortions, He et al. ~\cite{he2019progressive} propose to achieve a more accurate color transfer by leveraging  semantically-meaningful dense correspondence between images. 
One line of work ameliorates this problem by imposing additional constraints on the loss function.
Luan et al.~\cite{luan2017deep} observe that constraining the transformation to be locally affine in color space pushes the result towards photorealism.

PhotoWCT~\cite{li2018closed} imposes a similar constraint as a postprocessing step, while LST~\cite{li2019learning} appends a spatial propagation network~\cite{liu2017learning} after the main style transfer network to learn to preserve the desired affinity.
Similarily, Puy et al.~\cite{puy2019flexible} propose a flexible network to perform artistic style transfer, and applies postprocessing after each learned update for photorealistic content.
Compared to these ad hoc approaches, where the photorealism constraint is a soft penalty, our model directly predicts local affine transforms, guaranteeing that the constraint is satisfied.

Another line of recent work shows that matching the statistics of auto-encoders is an effective way to parameterize style transfer~\cite{huang2017arbitrary,li2017universal,li2018closed,li2019learning,yoo2019photorealistic}.
Moreover, they show that distortions can be reduced by preserving high frequencies using unpooling~\cite{li2018closed} or wavelet transform residuals~\cite{yoo2019photorealistic}.

Our work unifies these two lines of research.
Our network architecture builds upon HDRnet~\cite{gharbi2017deep}, which was first
employed in the context of learning image enhancement and tone manipulation.
Given a large dataset of input/output pairs, it learns local affine transforms that best reproduces the operator.
The network is small and the learned transforms that are intentionally constrained to be incapable of introducing artifacts such as noise or false edges.
These are exactly the properties we want and indeed, Gharbi et. al. demonstrated style transfer in their original paper.
However, when we applied HDRnet to our more diverse dataset, we found a number of artifacts (Figure~\ref{fig:inspiration}).
This is because HDRnet does not explicitly model style transfer and instead learns by memorizing what it sees during training and projecting the function onto local affine transforms.
Therefore, it will require a lot of training data and generalize poorly.
Since HDRnet learns local affine transforms from low-level image features, our strategy is to start with statistical feature matching using Adaptive Instance Normalization~\cite{huang2017arbitrary} to build a joint distribution. By explicitly modeling the style transformation as a distribution matching process, our network is capable of generalizing to unseen or adversarial inputs (Figure~\ref{fig:inspiration}).

\section{Method}

Our method is based on a single feed-forward deep neural network.
It takes as input two images, a \emph{content photo} $I_c$, and an arbitrary \emph{style image} $I_s$, producing a photorealistic output $O$ with the former's content but the latter's style.
Our network is ``universal''---after training on a diverse dataset of content/style pairs, it can generalize to novel input combinations.
Its architecture is centered around the core idea of learning local affine transformations, which inherently enforce the photorealism constraint.

\subsection{Background}

For completeness, we first summarize the key ideas of recent work. 

\paragraph{Content and Style.}
The Neural Style Transfer~\cite{gatys2016image} algorithm is based on an optimization that minimizes a loss balancing the output image's fidelity to the input images' content and style:

\begin{align}
\centering
\mathcal{L}_{g} &= \alpha \mathcal{L}_{c} + \beta \mathcal{L}_{s}~~~~~~~~~~~~~\mathrm{\ \ with\ \ } \label{eq:loss_gatys} \\
\mathcal{L}_{c} &= \sum_{i=1}^{N_c} \left\| F_{i}[O]-F_{i}[I_{c}] \right\|_{2}^{2} \mathrm{\ \ and\ \ }
\mathcal{L}_{s} = \sum_{i=1}^{N_s} \left\| G_{i}[O]-G_{i}[I_{s}] \right\|_{F}^{2},
\label{eq:loss_content_and_style}
\end{align}
where $N_c$ and $N_s$ denote the number of intermediate layers selected from a pretrained VGG-19 network~\cite{simonyan2014very} to represent image content and style, respectively. Scene content is captured by the feature maps $F_i$ of intermediate layers of the VGG network, and style is captured by their Gram matrices $G_i[\cdot] = F_i[\cdot]F_i[\cdot]^T$. $||\cdot||_F$ denotes the Frobenius norm.

\paragraph{Statistical Feature Matching.}
Instead of directly minimizing the loss in Equation~\ref{eq:loss_gatys}, followup work shows that it is more effective to match the statistics of feature maps at the bottleneck of an auto-encoder.
Variants of the whitening and coloring transform~\cite{li2017universal,li2018closed,yoo2019photorealistic} normalize the singular values of each channel, while Adaptive Instance Normalization (AdaIN)~\cite{huang2017arbitrary} proposes a simple scheme using the mean $\mu(\cdot)$ and the standard deviation $\sigma(\cdot)$ of each channel:
\begin{equation}
\label{eq:adain}
\mathrm{AdaIN}(x, y) = \sigma(y)\left(\frac{x-\mu(x)}{\sigma(x)}\right) + \mu(y),
\end{equation}
where $x$ and $y$ are content and style feature channels, respectively. Due to its simplicity and reduced cost, we also adopt AdaIN layers in our network architecture as well as its induced style loss~\cite{huang2017arbitrary,li2017demystifying}:
\begin{equation}
\label{eq:loss_style_adain}
\begin{split}
\mathcal{L}_{sa} = \sum_{i=1}^{N_S} \left\| \mu(F_i[O]) - \mu(F_i[I_s]) \right\|_{2}^{2} +
                   \sum_{i=1}^{N_S} \left\| \sigma(F_i[O]) - \sigma(F_i[I_s]) \right\|_{2}^{2}.
\end{split}
\end{equation}

\begin{figure}[t]
  %\centering
  \vspace{-0.2cm}
  \hspace{-0.5cm}
  \includegraphics[width=1.05\linewidth]{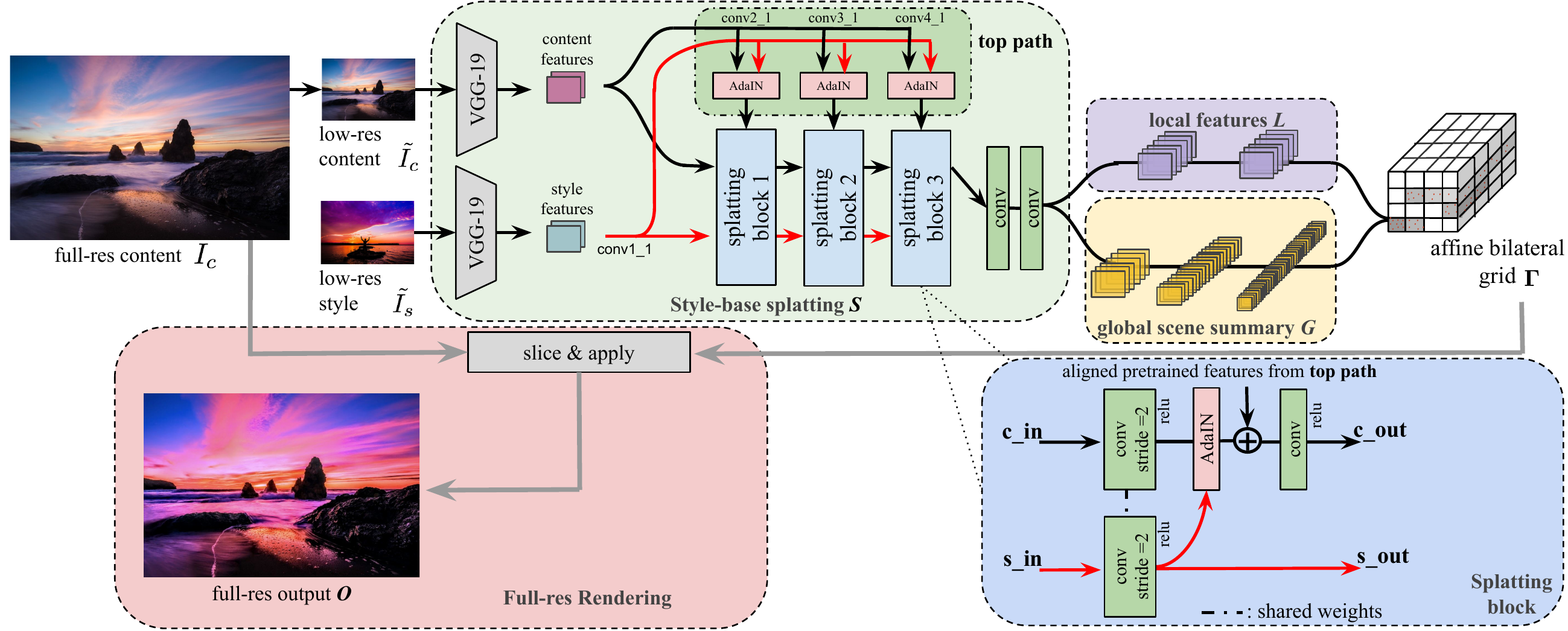}
  \vspace{-0.2cm}
  \caption{\textbf{Model architecture.}
  Our model starts with a low-resolution coefficient prediction stream that uses style-based splatting blocks \textbf{\textit{S}} to build a joint distribution between the low-level features of the input content/style pair. This distribution is fed to bilateral learning blocks \textbf{\textit{L}} and \textbf{\textit{G}} to predict an affine bilateral grid $\Gamma$.
  Rendering, which runs at full-resolution, performs the minimal per-pixel work of sampling from $\Gamma$ a $3 \times 4$ matrix and then multiplying.
  }
  \label{fig:architecture}
  \vspace{-0.4cm}
\end{figure}

\paragraph{Bilateral Space.}
Bilateral space was first introduced by Paris and Durand~\cite{paris2006fastapprox} in the context of fast edge-aware image filtering.
A 2D grayscale image $I(x,y)$ can be ``lifted'' into bilateral space as a sparse collection of 3D points $\{x_j, y_j, I_j\}$ in the augmented 3D space.
In this space, linear operations are inherently edge-aware because Euclidean distances preserve edges.
They prove that bilateral filtering is equivalent to \emph{splatting} the input onto a regular 3D \emph{bilateral grid}, blurring, and \emph{slicing} out the result using trilinear interpolation at the input coordinates $\{x_j, y_j, I_j\}$.
Since blurring and slicing are low-frequency operations, the grid can be low-resolution, dramatically accelerating the filter.

Bilateral Guided Upsampling (BGU)~\cite{chen2016bilateral} extends the bilateral grid to represent transformations between images.
By storing at each cell an affine transformation, an \emph{affine bilateral grid} can encode any image-to-image transformation given sufficient resolution.
The pipeline is similar: \emph{splat} both input and output images onto a bilateral grid, blur, and perform a per-pixel least squares fit.
To apply the transform, \emph{slice} out a per-pixel affine matrix and multiply by the input color.
BGU shows that this representation can accelerate a variety of imaging operators and that the approximation degrades gracefully with resolution when suitably regularized.
Affine bilateral grids are constrained to produce an output that is a smoothly varying, edge-ware, and locally affine transformation of the input.
Therefore, it fundamentally cannot produce false edges, amplify noise, and inherently obeys the photorealism constraint.

Gharbi et al.~\cite{gharbi2017deep} showed that slicing and applying an affine bilateral grid are sub-differentiable and therefore can be incorporated as a layer in a deep neural network and learned using gradient descent.
They demonstrated that their HDRnet architecture can effectively learn to reproduce many photographic tone mapping and detail manipulation tasks, regardless of whether they are algorithmic or artist-driven.

\subsection{Network Architecture}
\label{subsec:network_architecture}

Our end-to-end differentiable network % is inspired by HDRnet and 
consists of two streams. The \emph{coefficient prediction} stream takes as input reduced resolution content $\widetilde{I_c}$ and style $\widetilde{I_s}$ images, learns the joint distribution between their low-level features, and predicts an affine bilateral grid~$\Gamma$.
The \emph{rendering} stream, unmodified from HDRnet, operates at full-resolution.
At each pixel $(x, y, r, g, b)$, it uses a learned lookup table to compute a ``luma'' value $z = g(r,g,b)$, slices out $A = \Gamma(x/w,y/h,z/d)$ (using trilinear interpolation), and outputs $O = A * (r, g, b, 1)^T$.
By decoupling coefficient prediction resolution from that of rendering, our architecture offers a tradeoff between stylization quality and performance. Figure~\ref{fig:architecture} summarizes the entire network and we describe each block below.

\begin{table}[t]
\centering
\begin{tabular}{c | cccccccc | cc | cccccc | cc }
    & $S_1^1$ & $S_1^2$ & $S_2^1$& $S_2^2$ & $S_3^1$ & $S_3^2$ & $C^7$ & $C^8$ & $L^1$ & $L^2$ & $G^1$ & $G^2$ & $G^3$ & $G^4$ & $G^5$& $G^6$& $F$ & $\Gamma$ \\ \hline
type  & $c$  & $c$  & $c$  & $c$  & $c$  & $c$ & $c$  & $c$ & $c$ & $c$ & $c$ & $c$ & $f_{c}$ & $f_{c}$ & $f_{c}$ & $f_{c}$ & $c$ & $c$  \\
stride & 2  & 1  & 2  & 1  & 2  & 1 & 2 & 1 & 1 & 1 & 2 & 2 & -  & -  & - & - & 1 & 1  \\
size  & 128 & 128 & 64 & 64 & 32  & 32 & 16  & 16 & 16 & 16 & 8 & 4 & -  & -  & - & - & 16 & 16 \\
channels & 8  & 8  & 16  & 16 & 32  & 32 & 64  & 64 & 64 & 64 & 64 & 64 & 256 & 128 & 64 & 64 & 64 & 96 \\ \hline
\end{tabular}
\caption{Details of our network architecture. $S_j^i$ denotes the $i$-$th$ layer in the $j$-$th$ splatting block. We apply AdaIN after each $S_j^1$. $L^i$, $G^i$, $F$, and $\Gamma$ refer to local features, global features, fusion, and learned bilateral grid respectively. Local and global features are concatenated before fusion $F$. $c$ and $f_c$ denote convolutional and fully-connected layers, respectively. Convolutions are all $3 \times 3$ except $F$, where it is $1 \times 1$. }
\label{tbl:layers_specifics}
\vspace{-0.2cm}
\end{table}

\subsubsection{Style-based Splatting.}
We aim to first learn a multi-scale model of the joint distribution between content and style features, and from this distribution, predict an affine bilateral grid.
Rather than using strided convolutional layers to directly learn from pixel data, we follow recent work~\cite{johnson2016perceptual,li2018closed,huang2017arbitrary} and use a pretrained VGG-19 network to extract low-level features from both images at four scales ($conv1\_1$, $conv2\_1$, $conv3\_1$, and $conv4\_1$).
We process these multi-resolution feature maps with a sequence of \emph{splatting blocks} inspired by the StyleGAN architecture~\cite{karras2019style} (Figure~\ref{fig:architecture}).
Starting from the finest level, each splatting block applies a stride-2 \emph{weight-sharing} convolutional layer to both content and style features, halving spatial resolution while doubling the number of channels (see Table~\ref{tbl:layers_specifics}).
The shared-weight constraint crucially allows the following AdaIN layer to learn the joint content/style distribution without correspondence supervision.
Once the content feature map is rescaled, we append it to the similarly AdaIN-aligned feature maps from the pretrained VGG-19 layer of the same resolution.
Since the content feature map now contains more channels, we use a stride-1 convolutional layer to select the relevant channels between learned-and-normalized vs. pretrained-and-normalized features.

We use three splatting blocks in our architecture, corresponding to the finest-resolution layers of the selected VGG features.
While using additional splatting blocks is possible, they are too coarse and replacing them with standard stride-2 convolutions makes little difference in our experiments.
Since this component of the network effectively learns the relevant bilateral-space content features based on its corresponding style, it can be thought of as \emph{learned style-based splatting}.

%------------------------------------------------
\subsubsection{Joint Bilateral Learning.}
With aligned-to-style content features in bilateral space, we seek to learn an affine bilateral grid that encodes a transformation that locally captures style and is aware of scene semantics.
Like HDRnet, we split the network into two asymmetric paths: a fully-convolutional \emph{local path} that learns local color transforms and thereby sets the grid resolution, and a \emph{global path}, consisting of both convolutional and fully-connected layers, that learns a summary of the scene and helps spatially regularize the transforms.
The local path consists of two stride $1$ convolutional layers, keeping the spatial resolution and number of features constant.
This provides enough depth to learn local affine transforms without letting its receptive field grow too large (and thereby discarding any notion of spatial position).

As we aim to perform universal style transfer without any explicit notion of semantics (e.g., per-pixel masks provided by an external pretrained network), we use a small network to learn a global notion of scene category.
Our global path consists of two stride $2$ convolutional layers to further reduce resolution, followed by four fully-connected layers to produce a $64-$element vector ``summary''.
We append the summary at each $x, y$ spatial location output from the local path and use a $1 \times 1$ convolutional layer to reduce the final output to $96$ channels.
These $96$ channels can be reshaped into a $8$ ``luma bins'' that separate edges, each storing a $3 \times 4$ affine transform.
We use the ReLU activation after all but the final $1 \times 1$ fusion layer and zero-padding for all convolutional layers.

\subsection{Losses}
\label{subsec:losses}

Since our architecture is fully differentiable, we can simply define our loss function on the generated output. We augment the content and style fidelity losses of Huang et al.~\cite{huang2017arbitrary} with a novel \textbf{bilateral-space Laplacian regularizer}, similar to the one in~\cite{gupta2016monotonic}:

\begin{equation}
\label{eq:loss_total}
\mathcal{L} = \lambda_c \mathcal{L}_c + \lambda_{sa} \mathcal{L}_{sa} + \lambda_r \mathcal{L}_r,
\end{equation}
where $\mathcal{L}_c$ and $\mathcal{L}_{sa}$ are the content and style losses defined in Equations~\ref{eq:loss_content_and_style}~and~\ref{eq:loss_style_adain}, and
\begin{equation}
\label{eq:laplacian_reg}
\mathcal{L}_r(\Gamma) = \sum_s \sum_{t \in N(s)} ||\Gamma[s] - \Gamma[t]||_F^2,
\end{equation}
where $\Gamma[s]$ is one cell of the estimated bilateral grid, and $\Gamma[t]$ one of its neighbors.

The Laplacian regularizer penalizes differences between adjacent cells of the bilateral grid (indexed by $s$ and finite differences computed over its six-connected neighbors $N(s)$) and encourages the learned affine transforms to be smooth in both space and intensity~\cite{chen2016bilateral,gupta2016monotonic}.
As we show in our ablation study (Sec~\ref{subsec:ablation}), the Laplacian regularizer is necessary to prevent visible grid artifacts.

We set $\lambda_{c}=0.5$, $\lambda_{sa}=1$, and $\lambda_{r}=0.15$ in all experiments.

\subsection{Training}
\label{subsec:training}

We trained our model on high-quality photos using Tensorflow~\cite{abadi2016tensorflow}, without any explicit notion of semantics. We use the Adam optimizer~\cite{kingma2015adam} with hyperparameters $\alpha=10^{-4}, \beta_1=0.9, \beta_2=0.999, \epsilon=10^{-8}$, and a batch size of $12$ content/style pairs.
For each epoch, we randomly split the data into 50000 content/style pairs. The training resolution is $256 \times 256$ and we train for a fixed $25$ epochs, taking two days on a single NVIDIA Tesla V100 GPU with 16 GB RAM.
Once the model is trained, inference can be performed at arbitrary resolution (since the bilateral grid can be scaled). To significantly reduce training time, we train the network at a fairly low resolution. As shown in Figure~\ref{fig:hd_4k}, the trained network still performs well even with 12 megapixel inputs. We attribute this to the fact that our losses are derived from pretrained VGG features, which are relatively invariant with respect to resolution.

%--------------------------------------------------------------------------------------------------------

\section{Results}
\label{sec:results}

For evaluation, we collected a test set of 400 high-quality images from websites. We compared our algorithm to the state of the art in photorealistic style transfer, and conducted a user study. Furthermore, we perform a set of ablation studies to better understand the contribution of various components.

Detailed comparisons with high-resolution images are included in the supplement.

\subsection{Ablation Studies}
\label{subsec:ablation}

\begin{figure}[ht]
\centering
% \begin{minipage}[t]{1.0\textwidth}
    \begin{tabular}{c@{\hspace{0.002\linewidth}}c@{\hspace{0.005\linewidth}}c@{\hspace{0.002\linewidth}}c@{\hspace{0.002\linewidth}}c@{\hspace{0.002\linewidth}}}
    
  \includegraphics[width = .12\linewidth,height=.14\linewidth]{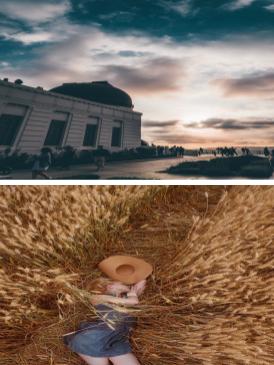} &
  \includegraphics[width = .2\linewidth,height=.14\linewidth]{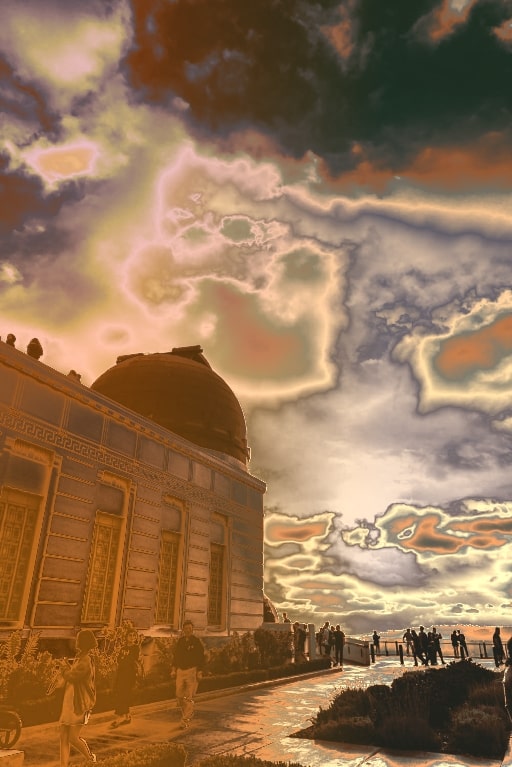} &
  \includegraphics[width = .2\linewidth,height=.14\linewidth]{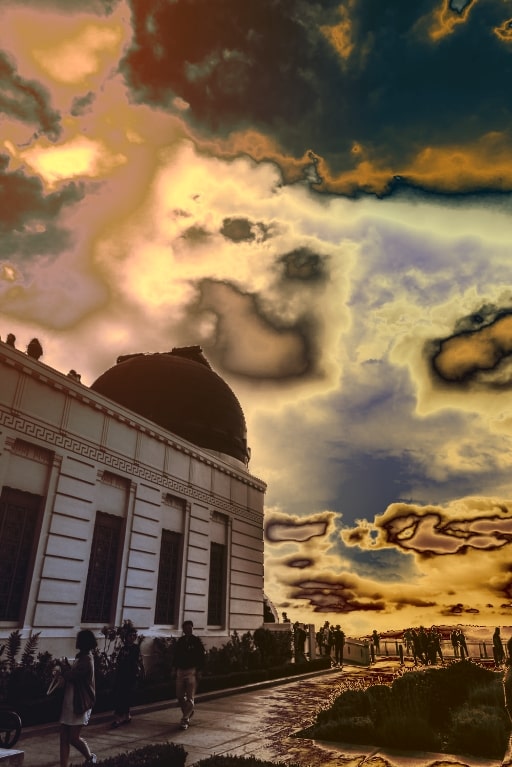} &
  \includegraphics[width = .2\linewidth,height=.14\linewidth]{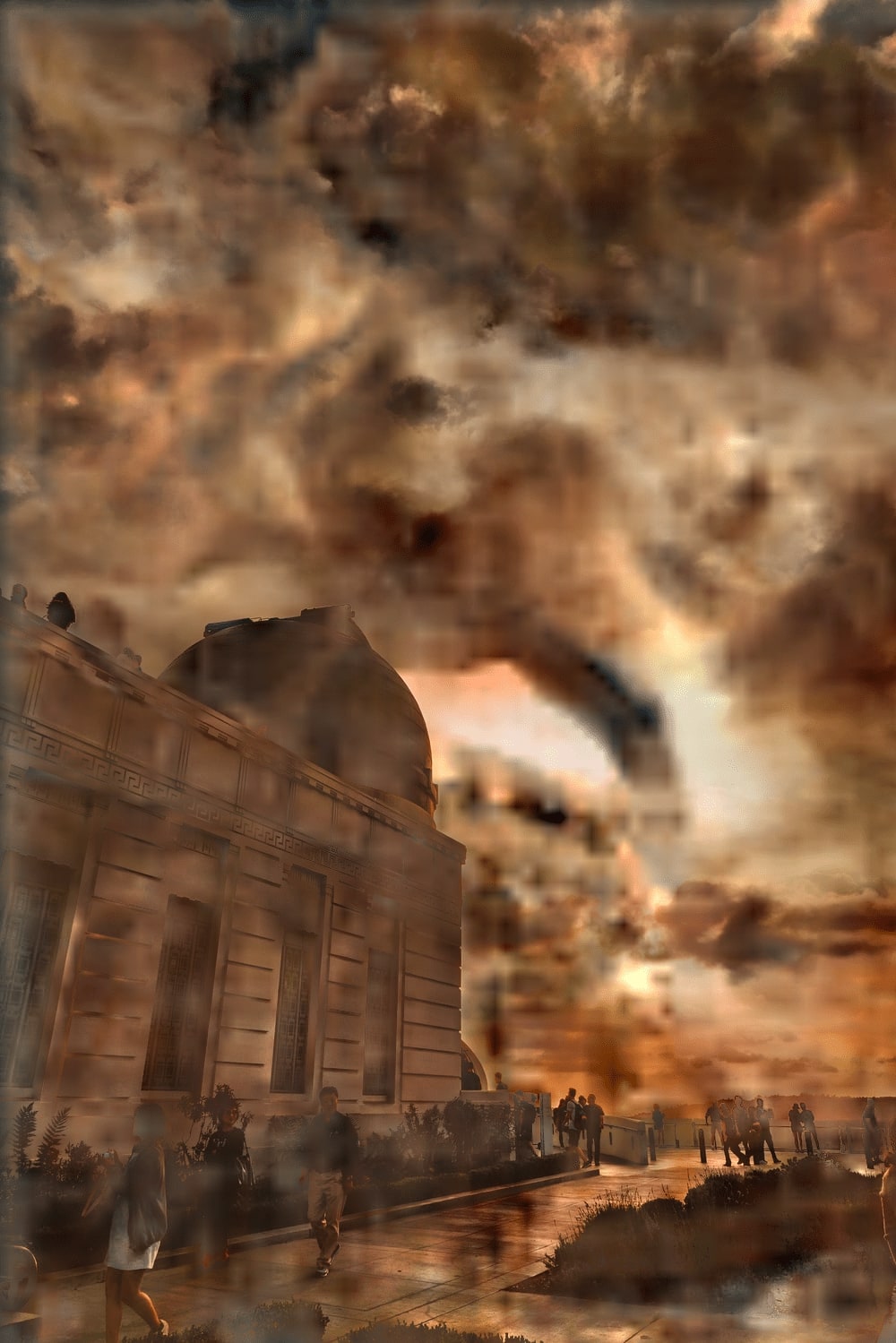} &
  \includegraphics[width = .2\linewidth,height=.14\linewidth]{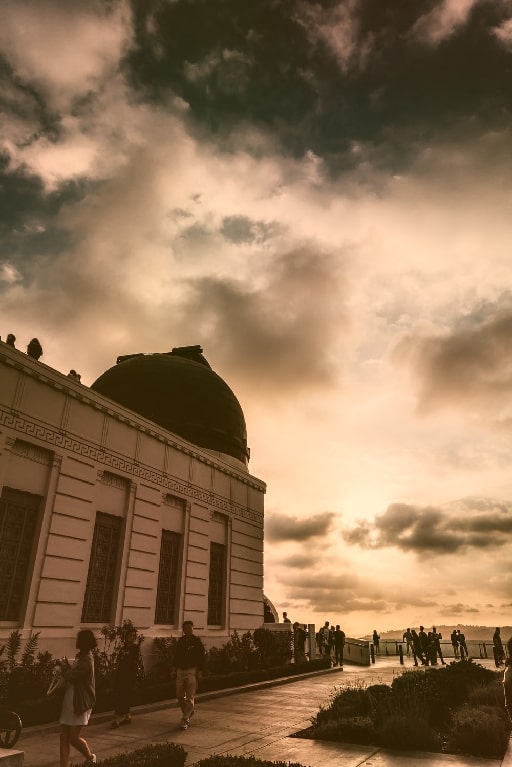} \\

  \includegraphics[width = .12\linewidth,height=.14\linewidth]{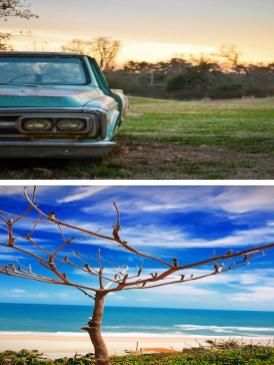} &
  \includegraphics[width = .2\linewidth,height=.14\linewidth]{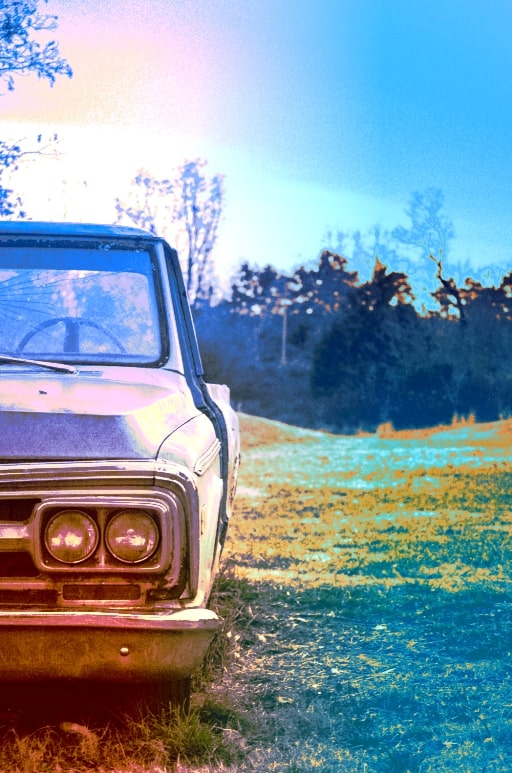} &
  \includegraphics[width = .2\linewidth,height=.14\linewidth]{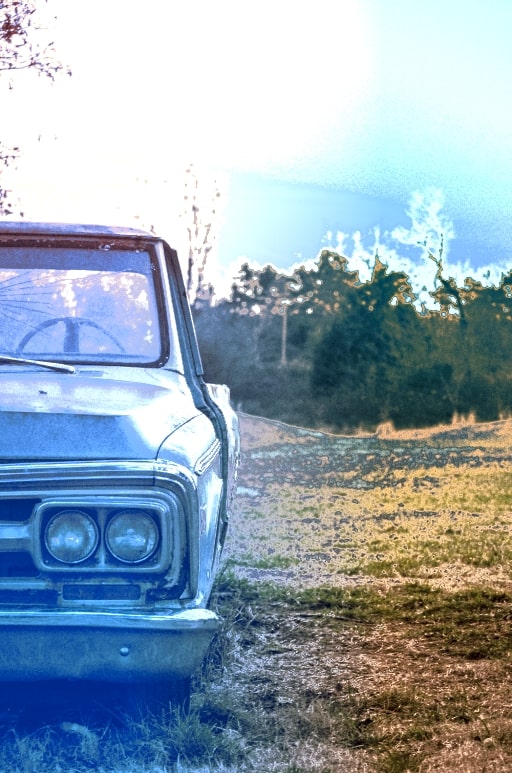} &
  \includegraphics[width = .2\linewidth,height=.14\linewidth]{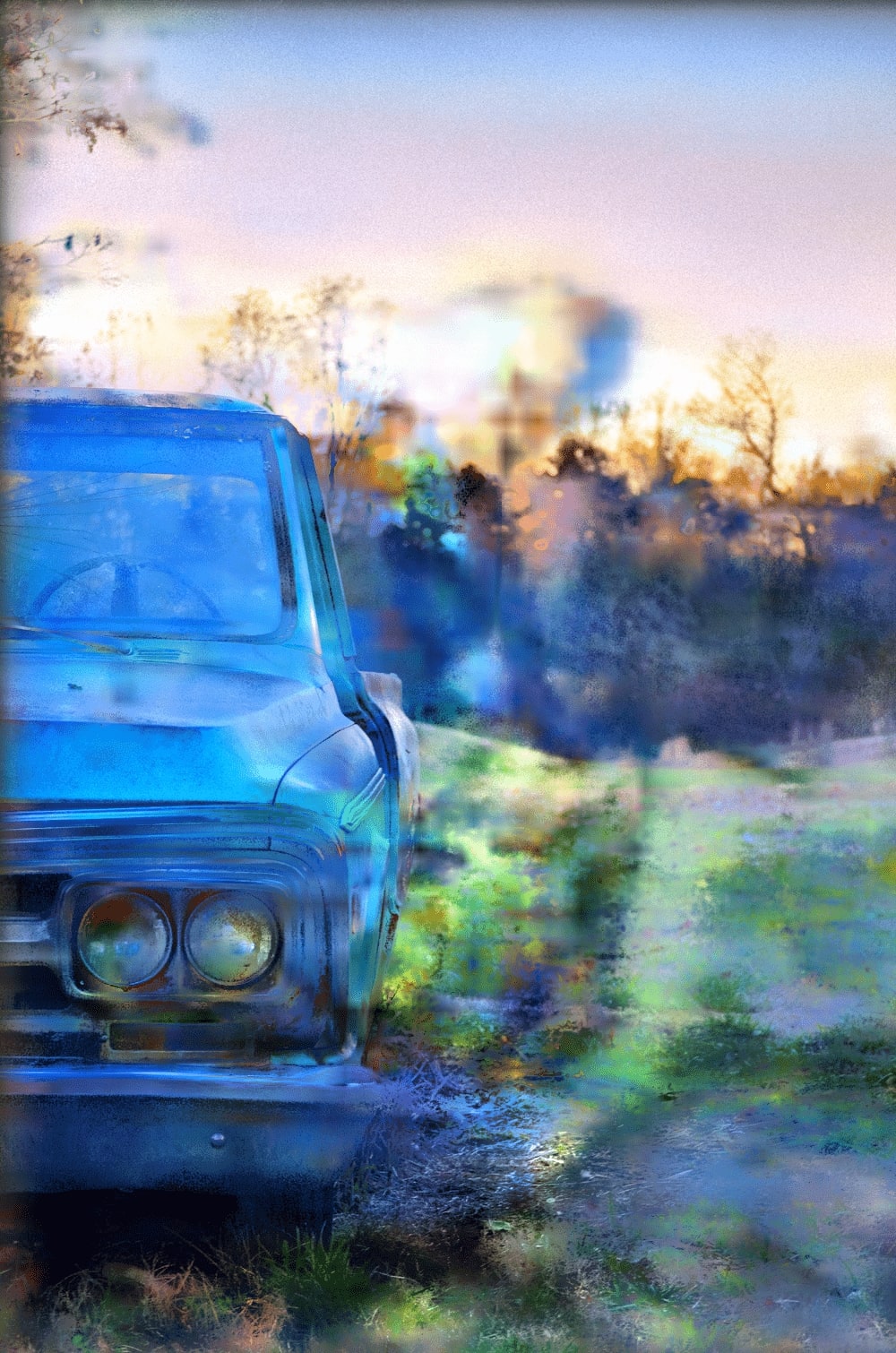} &
  \includegraphics[width = .2\linewidth,height=.14\linewidth]{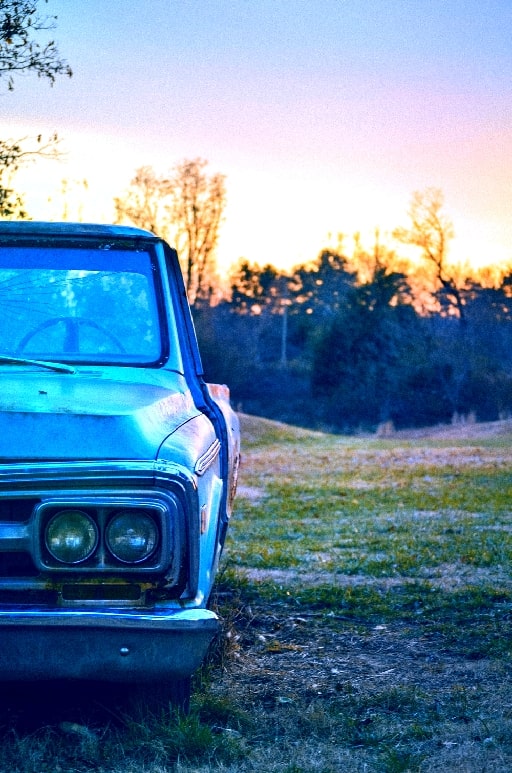} \\
  { (a) Inputs } & { (b) AdaIN $\rightarrow$ grid } & { (c) WCT $\rightarrow$ grid } & { (d) AdaIN+BGU } & { (e) Ours } \\

  \includegraphics[width = .12\linewidth,height=.14\linewidth]{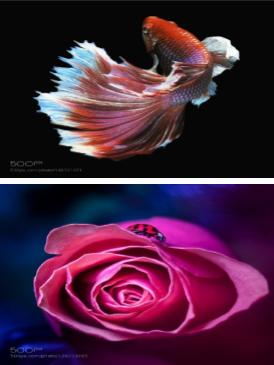} &
  \includegraphics[width = .2\linewidth,height=.14\linewidth]{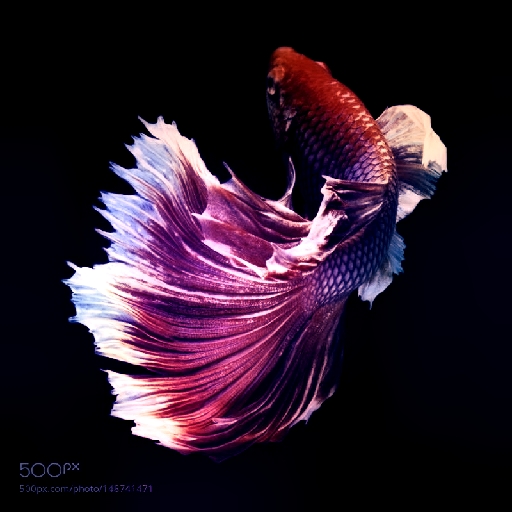} &
  \includegraphics[width = .2\linewidth,height=.14\linewidth]{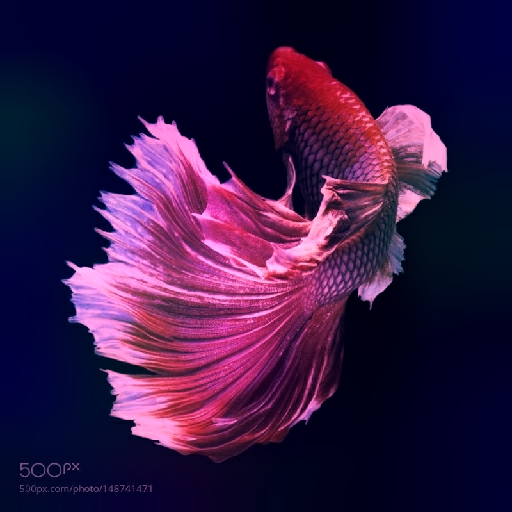} &
  \includegraphics[width = .2\linewidth,height=.14\linewidth]{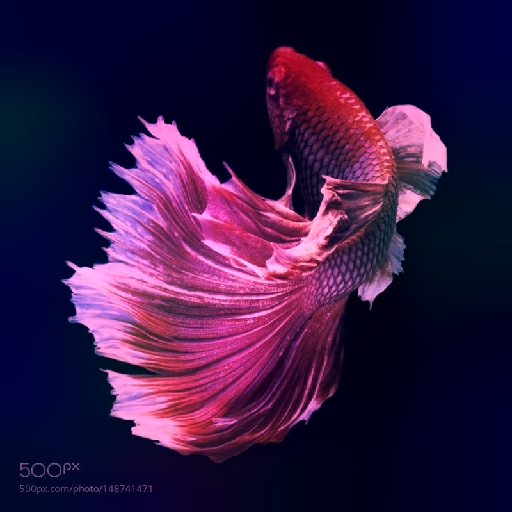} &
  \includegraphics[width = .2\linewidth,height=.14\linewidth]{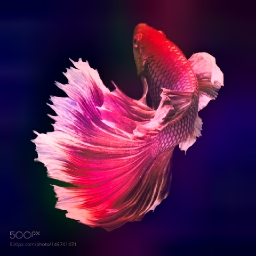}\\

  \includegraphics[width = .12\linewidth,height=.14\linewidth]{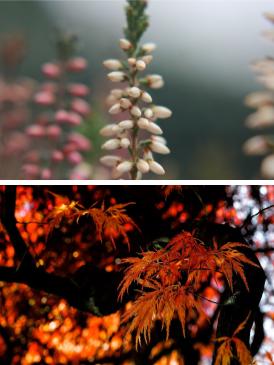} &
  \includegraphics[width = .2\linewidth,height=.14\linewidth]{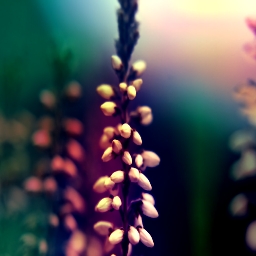} &
  \includegraphics[width = .2\linewidth,height=.14\linewidth]{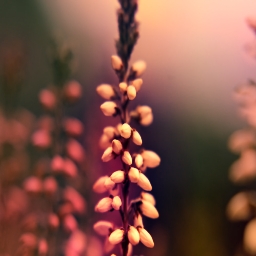} &
  \includegraphics[width = .2\linewidth,height=.14\linewidth]{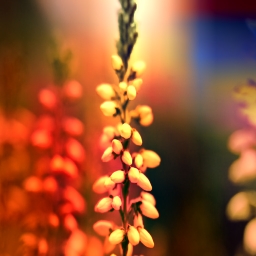} &
  \includegraphics[width = .2\linewidth,height=.14\linewidth]{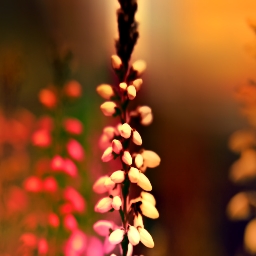}\\
  
  { (f) Inputs }& { (g) Block1 } &{(h) Block2}& { (i) Block3 }& { (j) Full results } \\
\end{tabular}
% \end{minipage}
\caption{\textbf{Ablation studies on splatting blocks.} (a)-(e): We demonstrate the importance of our splatting architecture by replacing it with baseline networks. (f)-(j): Visualization of the contribution of each splatting block by disabling statistical feature matching on the others.}
\label{fig:splat_viz}
  \vspace{-0.1cm}
\end{figure}

\subsubsection{Style-based Splatting Design.}
We conduct multiple ablations to show the importance of our style-based splatting blocks \textbf{\textit{S}}.

First, we consider replacing \textbf{\textit{S}} with two baseline networks: AdaIN~\cite{huang2017arbitrary} or WCT~\cite{li2017universal}.
Starting with the same features extracted from VGG-19, we perform feature matching using AdaIN or WCT.
The rest of the network is unchanged: that is, we attempt to learn local and global features directly from the baseline encoders and predict affine bilateral grids.
The results in Figure~\ref{fig:splat_viz} (b) and (c) show that while content is preserved, there is both an overall color cast as well as inconsistent blotches.
The low-resolution features simply lack the information density to learn even global color correction.

Second, to illustrate the contribution of each splatting block, we visualize our network's output when all but one block is disabled (including the top path inputs).
As shown in Figure~\ref{fig:splat_viz}(f--j), earlier, finer resolution blocks learn texture and local contrast, while later blocks capture more global information such as the style input's dominant color tone, which is consistent with our intuition. By combining all splatting blocks at three different resolutions, our model merges these features at multiple scales into a joint distribution.

%--------------------------------------------------------------------------------------------------------

\subsubsection{Network component ablations.}

\begin{figure}[t]
\centering
\begin{tabular}{c@{\hspace{0.005\linewidth}}c@{\hspace{0.005\linewidth}}c@{\hspace{0.005\linewidth}}c@{\hspace{0.005\linewidth}}c@{\hspace{0.005\linewidth}}c@{\hspace{0.005\linewidth}}c@{\hspace{0.005\linewidth}}c}

  \includegraphics[width = .12\linewidth,height=.14\linewidth]{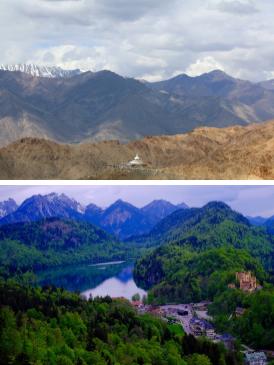} &
  \includegraphics[width = .2\linewidth,height=.14\linewidth]{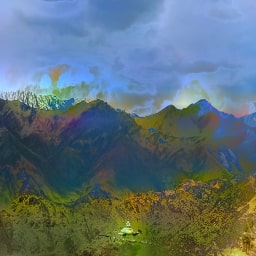} &
  \includegraphics[width = .2\linewidth,height=.14\linewidth]{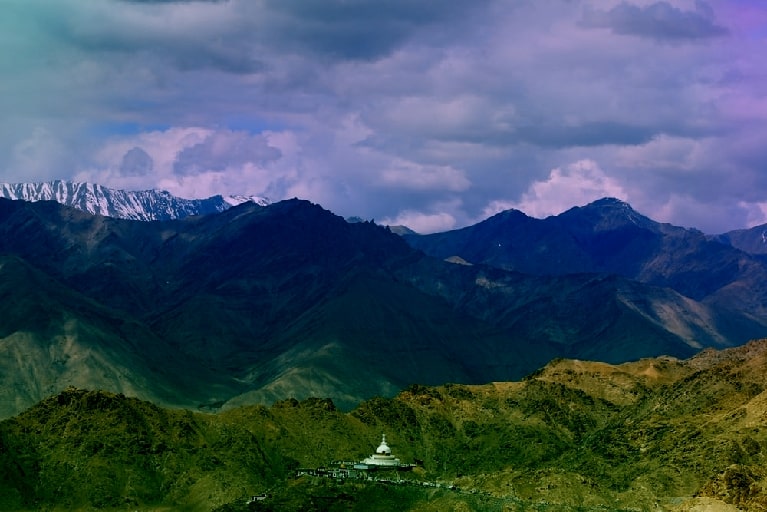} &
  \includegraphics[width = .2\linewidth,height=.14\linewidth]{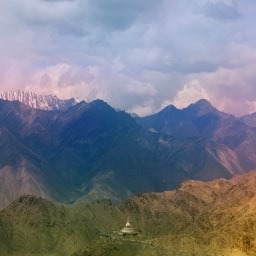} &
  \includegraphics[width = .2\linewidth,height=.14\linewidth]{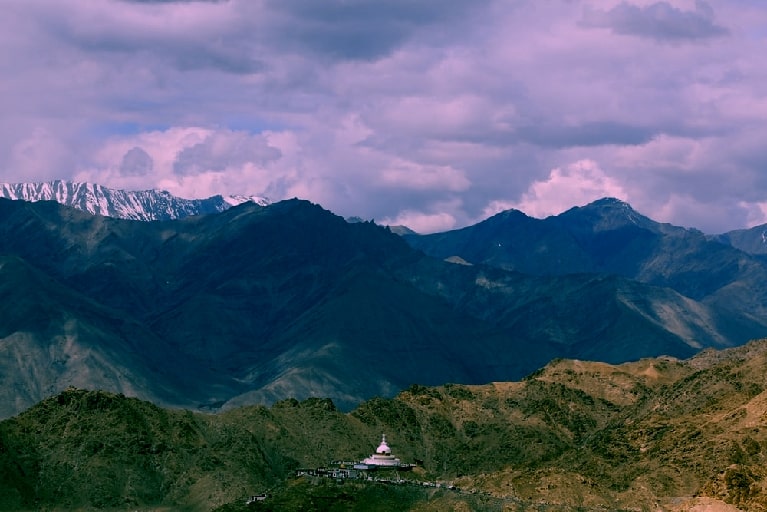} & \\

  \includegraphics[width = .12\linewidth,height=.14\linewidth]{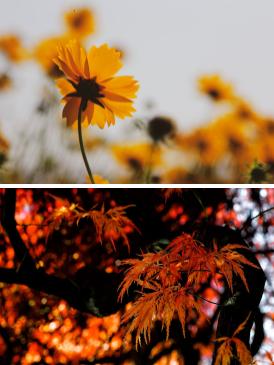} &
  \includegraphics[width = .2\linewidth,height=.14\linewidth]{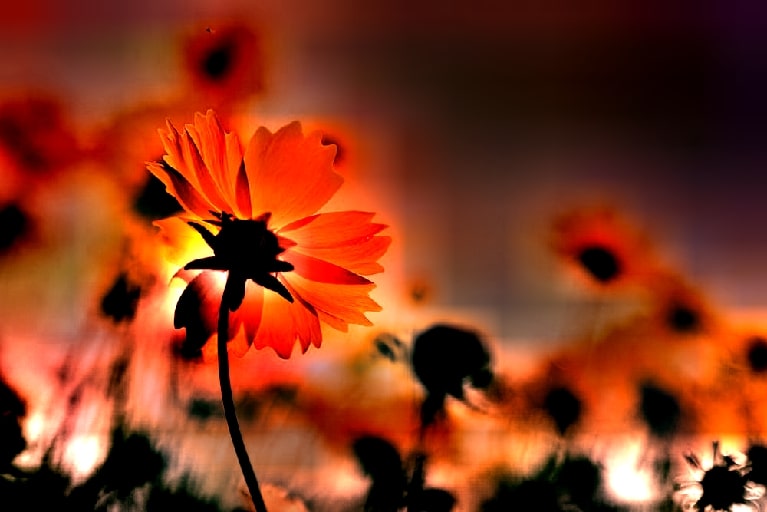} &
  \includegraphics[width = .2\linewidth,height=.14\linewidth]{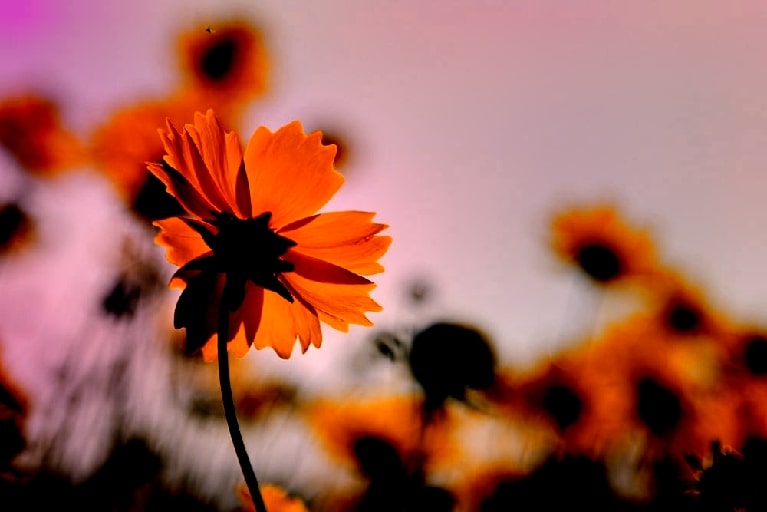} &
  \includegraphics[width = .2\linewidth,height=.14\linewidth]{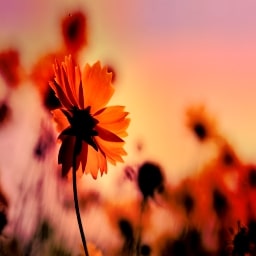} &
  \includegraphics[width = .2\linewidth,height=.14\linewidth]{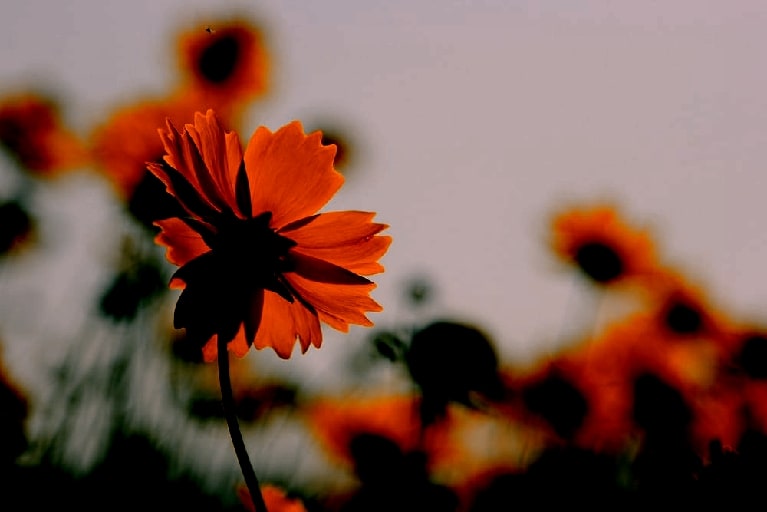} &\\
  
  (a) Inputs & (b) No $ \mathcal{L}_r$ & (c) No summary & (d) No top path & (e) Full results
   %inputs &{ No $\mathcal{L}_r$ } &  { No Global } & { Ours }  \\
\end{tabular}

\caption{Network component ablations.}
\label{fig:ablation_components}
%\vspace{-1.em}
\end{figure}

To demonstrate the importance of other blocks of our network, in Figure~\ref{fig:ablation_components}, we further compare our network with three variants: one trained without the bilateral-space Laplacian regularization loss (Equation~\ref{eq:loss_style_adain}), one without the global scene summary (Figure~\ref{fig:architecture}, yellow block), and one without ``top path'' inputs (Figure~\ref{fig:architecture}, dark green block).
We also show that our network learns stylization parameterized as local affine transforms.

Figure~\ref{fig:ablation_components} (b) shows distinctive dark halos when bilateral-space Laplacian regularization is absent. This is due to the fact that the network can learn to set regions of the bilateral grid to zero where it does not encounter image data (because images occupy a sparse 2D manifold in the grid's 3D domain). When sliced, the result is a smooth transition between black and the proper transform.

In Figure~\ref{fig:ablation_components}(c), it shows the global summary helps with spatial consistency. For example, in \emph{mountain} photo, the left part of sky is saturated while the right part of mountain is slightly washed out, while the output of our full network in Figure~\ref{fig:ablation_components}(e) has more spatially consistent color. This is consistent with the observation in Gharbi et al.~\cite{gharbi2017deep}. 

% TOP path:
Figure~\ref{fig:ablation_components}(d), demonstrates that selecting between learned-and-normalized vs. pretrained-and-normalized features (Figure~\ref{fig:architecture}, ``top path'') is also necessary. The results show distinctive patches of incorrect color characteristic of the network locally overfitting to the style input. Adaptively selecting between learned and pretrained features at multiple resolutions eliminates this inconsistency.

% AdaIN+BGU
Finally, we also show that our network learns stylization parameterized as local affine transforms and not a simple edge-aware interpolation. We run the full AdaIN network~\cite{huang2017arbitrary} on our $256 \times 256$ content and style images to produce a low-resolution stylized result. We then use BGU~\cite{chen2016bilateral} to fit a $16 \times 16 \times 8$ affine bilateral grid (the same resolution as our network) and slice it with the full-resolution input to produce a full-resolution output. Figure~\ref{fig:splat_viz} (d) shows that this strategy works quite poorly: since AdaIN's output exhibits spatial distortions even at $256 \times 256$, there is no affine bilateral grid for BGU to find that can fix them.

\begin{figure}[t!]
\begin{minipage}[t]{0.46\textwidth}
% \hspace*{-0.5cm}
    \begin{tabular}{c@{\hspace{0.005\linewidth}}c@{\hspace{0.005\linewidth}}c@{\hspace{0.005\linewidth}}c@{\hspace{0.005\linewidth}}c}
      \includegraphics[width = .235\linewidth,height=.18\linewidth]{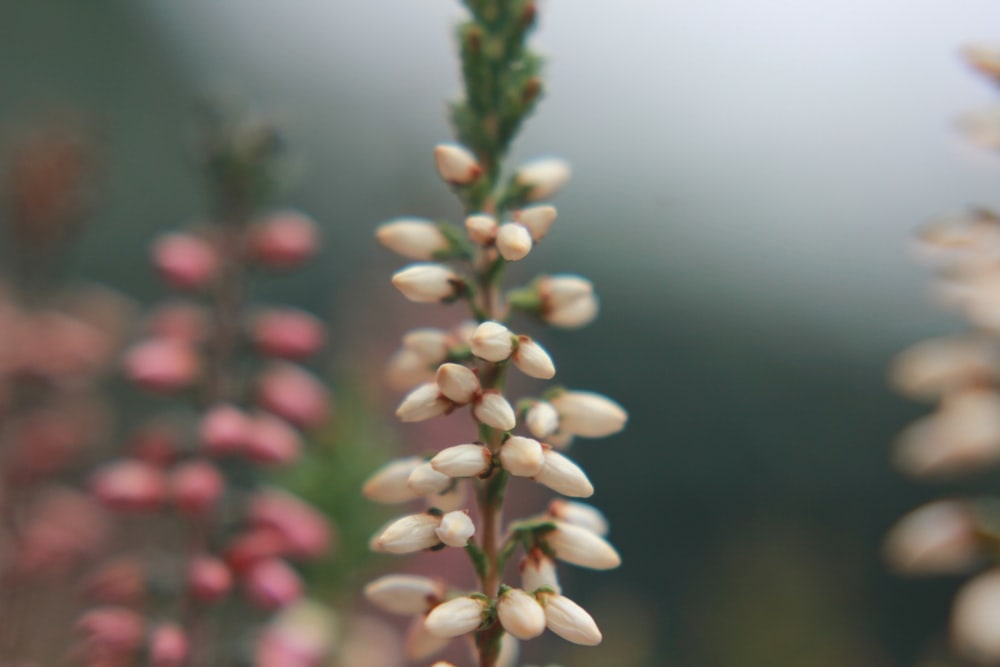} &
      \includegraphics[width = .235\linewidth,height=.18\linewidth]{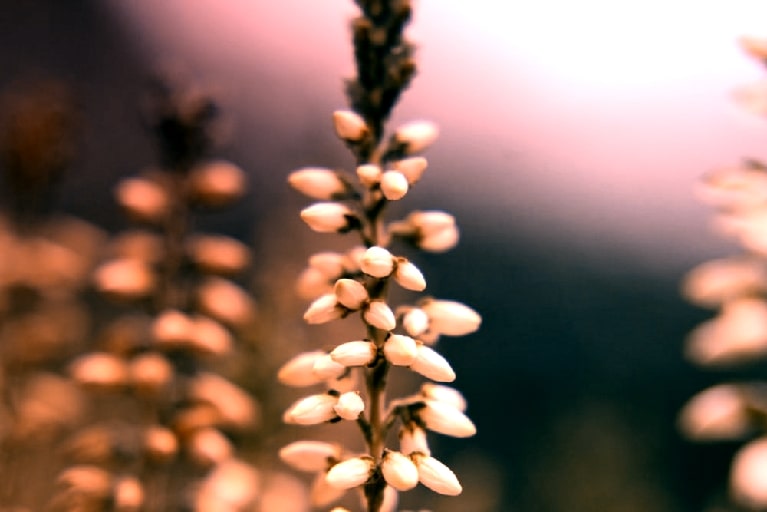} &
      \includegraphics[width = .235\linewidth,height=.18\linewidth]{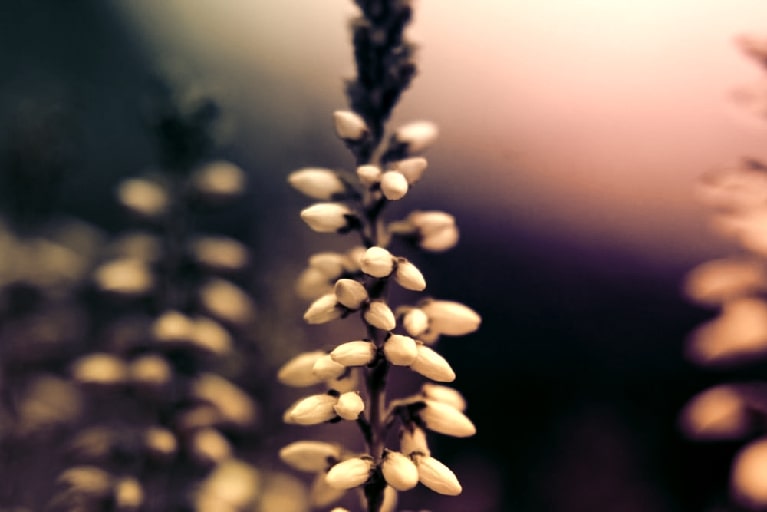} &
      \includegraphics[width = .235\linewidth,height=.18\linewidth]{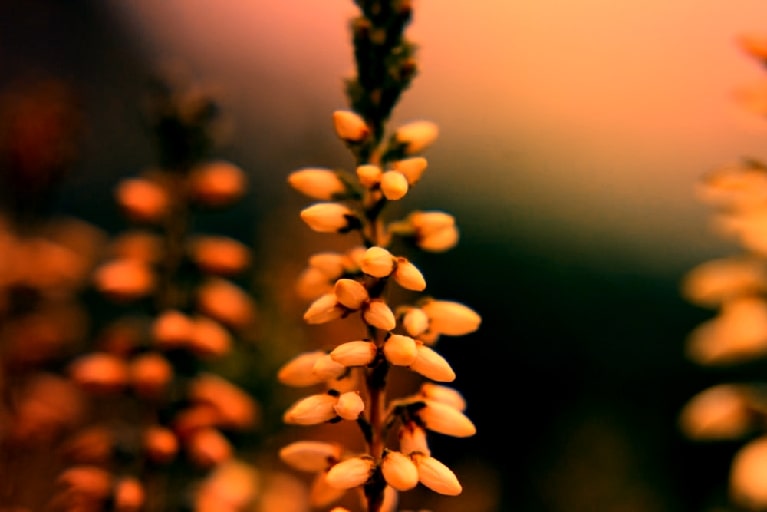} & \\
      Content & 1x1x8 & 2x2x8 & 8x8x8 & \\
      \includegraphics[width = .235\linewidth,height=.18\linewidth]{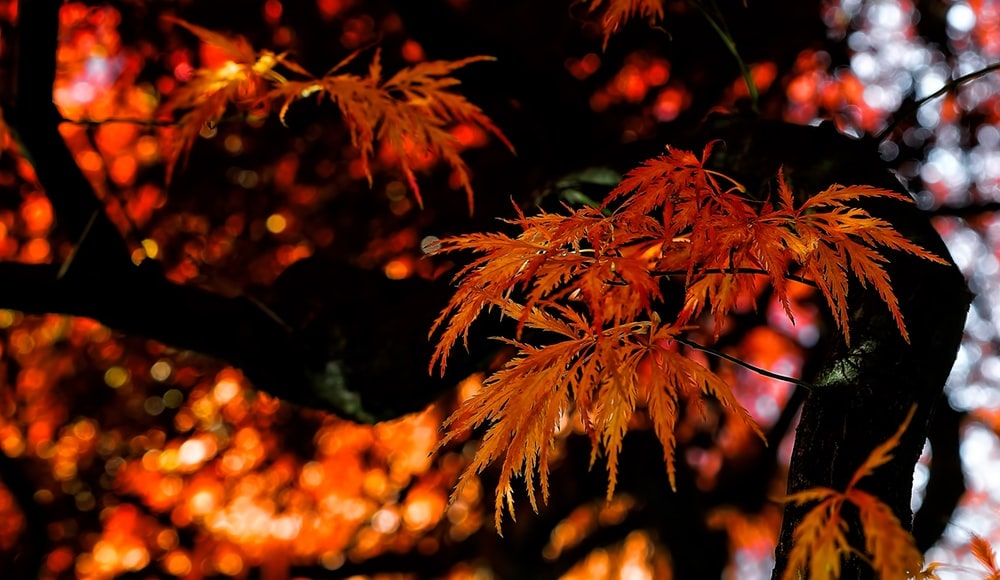} &
      \includegraphics[width = .235\linewidth,height=.18\linewidth]{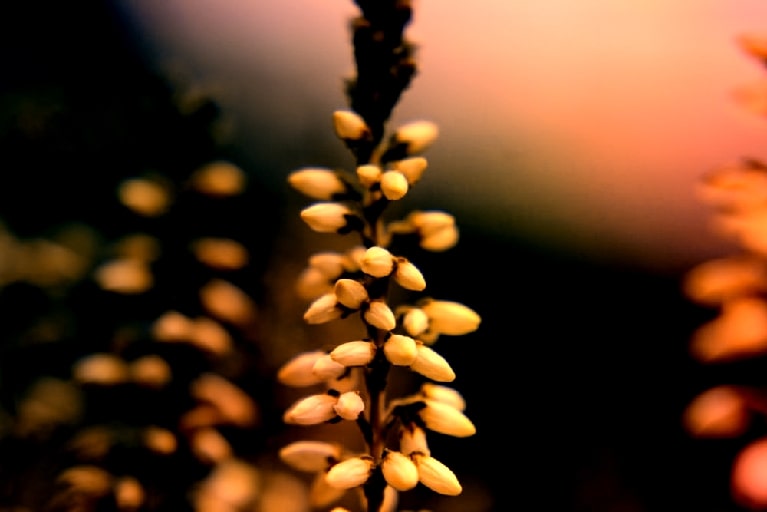} &
      \includegraphics[width = .235\linewidth,height=.18\linewidth]{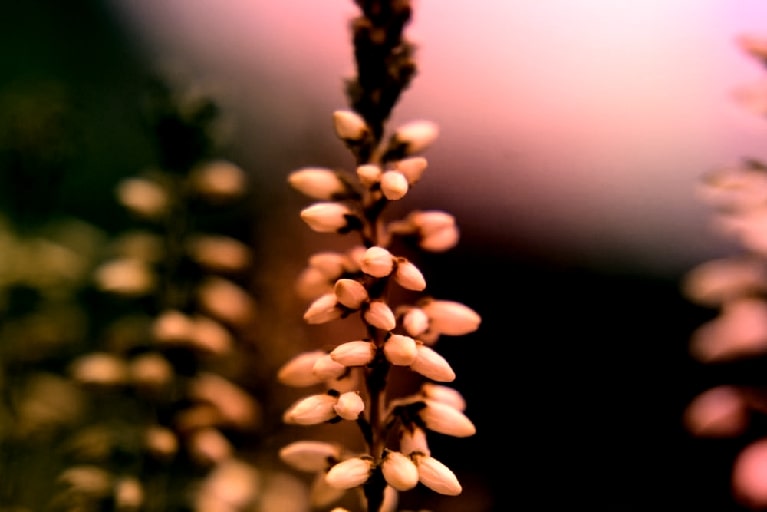} &
      \includegraphics[width = .235\linewidth,height=.18\linewidth]{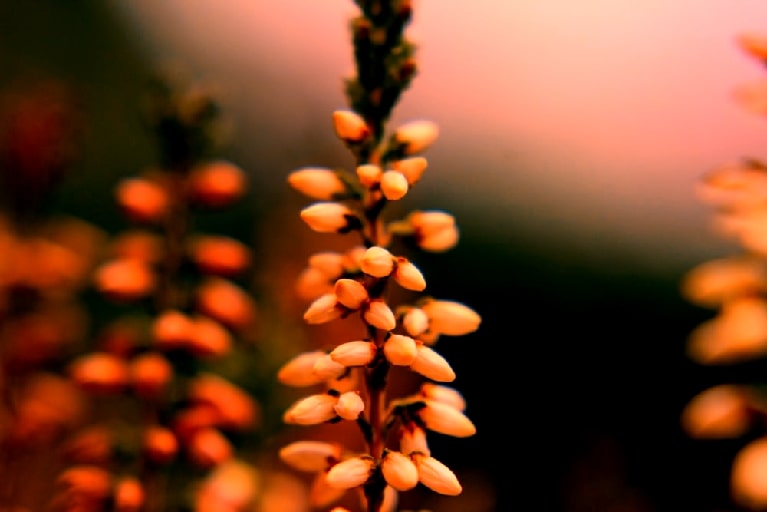} & \\
      Style & 16x16x1 & 16x16x2 & 16x16x8 & \\
    \end{tabular}
\vspace{-0.24em}
\centering\caption{Output using grids with different spatial (top) or luma resolutions (bottom) ($\mathrm{w} \times \mathrm{h} \times \mathrm{luma\ bins}$).}
\label{fig:grid_res}
\end{minipage}
\hfill
\begin{minipage}[t]{0.52\textwidth}
    \begin{tabular}{c@{\hspace{0.005\linewidth}}c@{\hspace{0.005\linewidth}}c@{\hspace{0.005\linewidth}}c@{\hspace{0.005\linewidth}}c}
    
      \includegraphics[width = .12\linewidth,height=.18\linewidth]{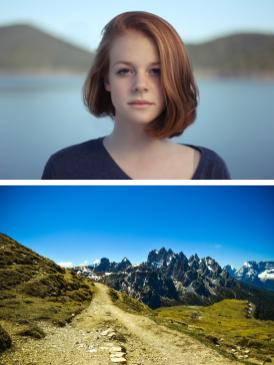} &
      \includegraphics[width = .27\linewidth,height=.18\linewidth]{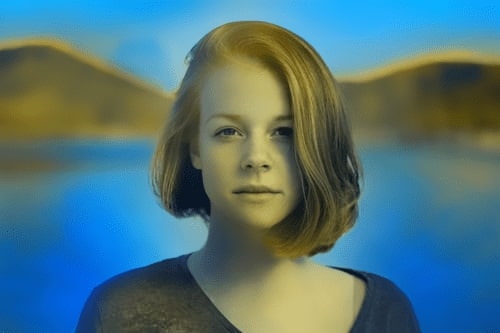} &
      \includegraphics[width = .27\linewidth,height=.18\linewidth]{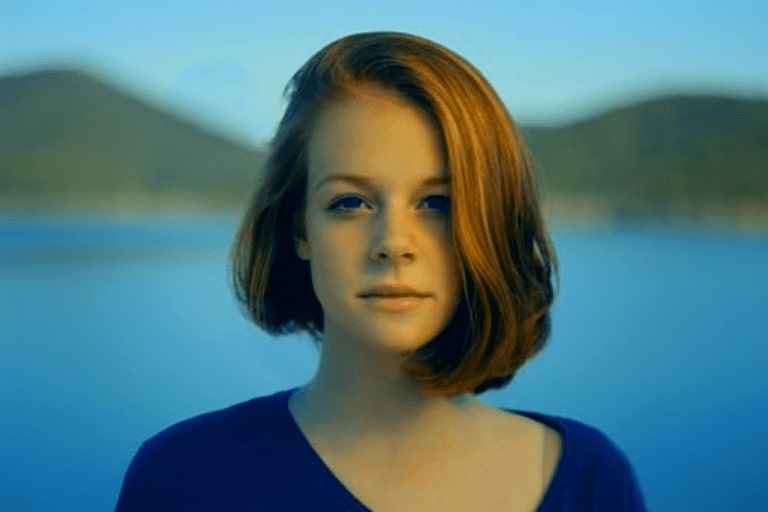} &
      \includegraphics[width = .27\linewidth,height=.18\linewidth]{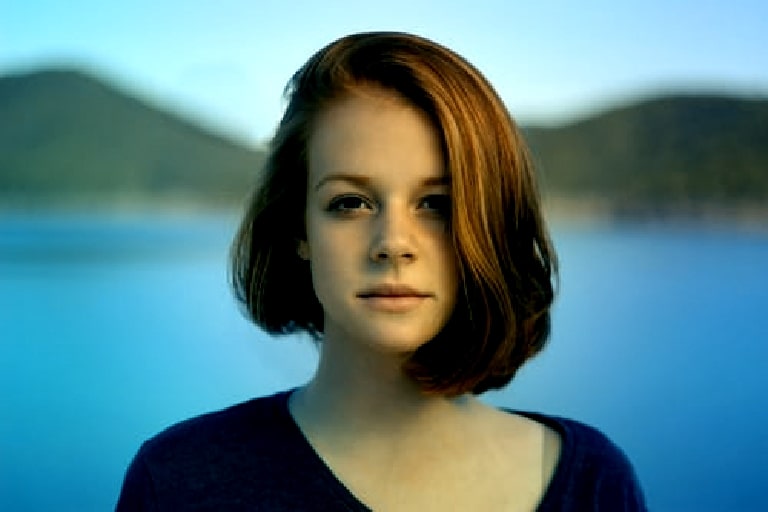} &\\
      
      \includegraphics[width = .12\linewidth,height=.18\linewidth]{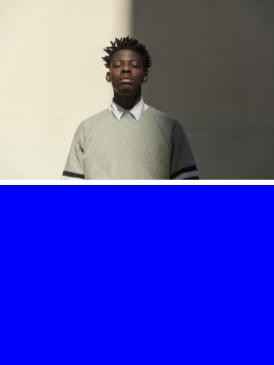} &
      \includegraphics[width = .27\linewidth,height=.18\linewidth]{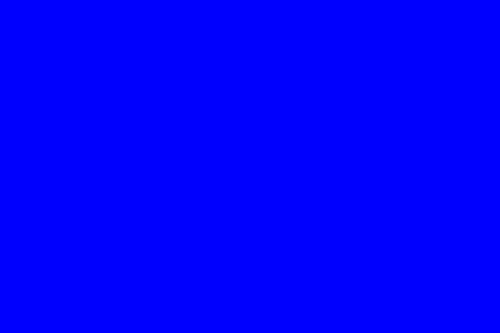} &
      \includegraphics[width = .27\linewidth,height=.18\linewidth]{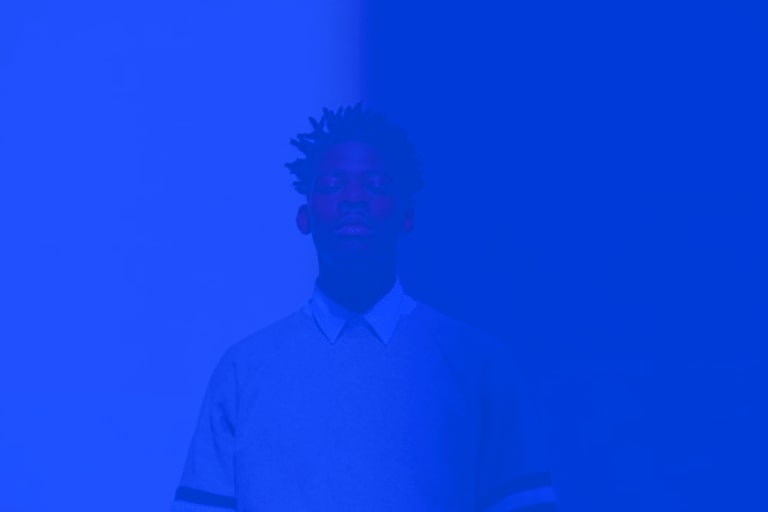} &
      \includegraphics[width = .27\linewidth,height=.18\linewidth]{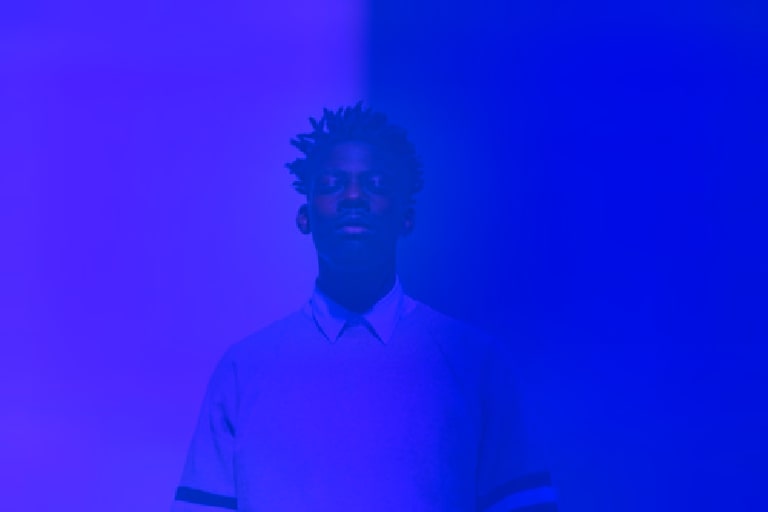} &\\
    
      inputs & PhotoWCT & WCT\textsuperscript{2} & Ours\\
    \end{tabular}
\vspace{-0.1em}    
\centering\caption{Our method is robust to adversarial inputs such as when the content image is a portrait (an unseen category) or even a monochromatic ``style''.}
\label{fig:extreme}
\end{minipage}

\end{figure}

\subsubsection{Grid Spatial Resolution.}
Figure~\ref{fig:grid_res} (top) shows how the spatial resolution of the grid affects stylization quality.
By fixing the number of luma bins at $8$, the $1 \times 1$ case is a single global curve, where the network learns an incorrectly colored compromise.
Going up to $2 \times 2$, the network attempts to spatially vary the transformation, with slightly different colors applied to different regions, but the result is still an unsatisfying tradeoff.
At $8 \times 8$, there is sufficient spatial resolution to yield a satisfying stylization result.

\subsubsection{Grid Luma Resolution.}
Figure~\ref{fig:grid_res} (bottom) also shows how the ``luma'' resolution affects stylization quality, with a fixed spatial resolution $16 \times 16$.
With $1$ luma bin, the network is restricted to predicting a single affine transform per tile.
Interpolating between $2$ luma bins reduces to a quadratic spline per tile, which is still insufficient for this image.
In our experiments, $8$ luma bins is sufficient for most images in our test set.

%---------------------------------------------------------------------------------

\subsection{Qualitative Results}

%-------------------------------------------------------------------

\begin{figure}[t!]
\vspace{-0.1cm}
\begin{minipage}[b]{0.51\textwidth}
\includegraphics[width = 0.97\columnwidth]{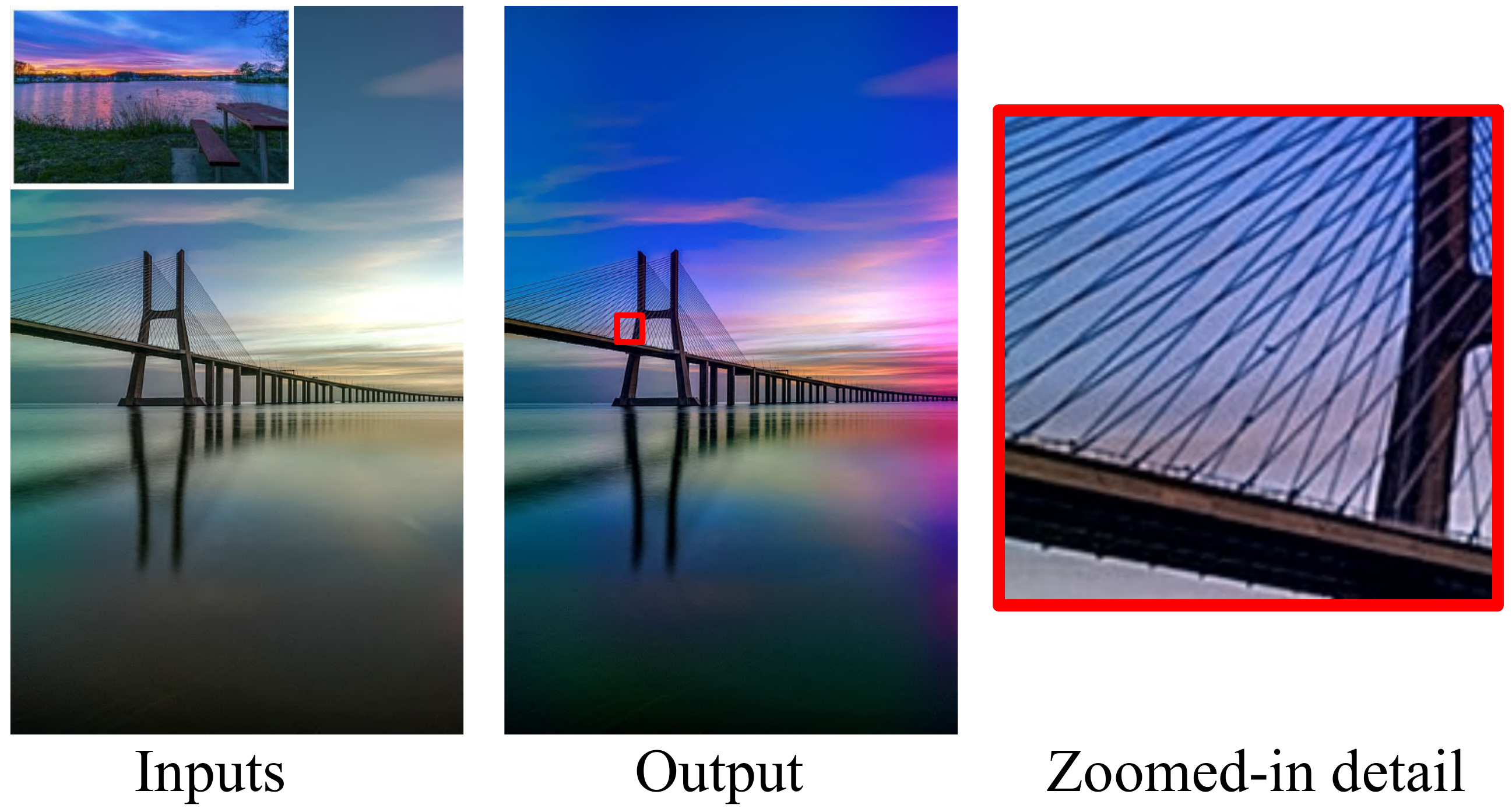} 
\centering
\vspace{-0.1cm}
\captionsetup{labelformat=empty}
\caption{(a) Output at 12 megapixels.}
\label{fig:hd_4k}
\end{minipage}
\hfill
\begin{minipage}[b]{0.49\textwidth}
\begin{minipage}[b]{1\textwidth}
%\hspace*{-1cm}
\scalebox{0.8}{\begin{tabular}{c | c | c  | c | c}
    Image Size         & PhotoWCT & LST & WCT\textsuperscript{2} & \textbf{Ours} \\ \hline
 %    256 $\times$ 256  & 428.22     & 0.47     & \textbf{0.07} & 3.31  & 0.29 \\
     512 $\times$ 512  & 0.68s     & 0.25s & 3.85s  & $\pmb{<}$ \textbf{5 ms} \\
    1024 $\times$ 1024 & 1.51s     & 0.84s & 6.13s  & $\pmb{<}$ \textbf{5 ms} \\
    1000 $\times$ 2000 & 2.75s     & OOM   & 10.94s & $\pmb{<}$ \textbf{5 ms} \\
    2000 $\times$ 2000 & OOM       & OOM   & OOM    & $\pmb{<}$ \textbf{5 ms} \\
    3000 $\times$ 4000 & OOM       & OOM   & OOM    & $\pmb{<}$ \textbf{5 ms} \\ \hline
\end{tabular}}
\centering
\captionsetup{labelformat=empty}
\vspace{-1em}
\caption{(b) Runtime.}
\end{minipage}

\vspace{0.2cm}
\begin{minipage}[b]{1\textwidth}
%\vspace{-1.5cm}
%\hspace*{-1cm}  
\scalebox{0.9}{\begin{tabular}{l |c | c  | c | c}
Mean Score & PhotoWCT  & LST & WCT\textsuperscript{2} & \textbf{Ours} \\ \hline
Photorealism     & 2.02    & 2.89    & \textbf{4.21}     & 4.14  \\
Stylization & 3.10    & 3.19   & 3.24     & \textbf{3.49} \\
Overall quality & 2.23   & 2.84    & 3.60     & \textbf{3.79} \\ \hline
\end{tabular}}
\vspace{-1em}
\centering
\captionsetup{labelformat=empty}
\caption{(c) User study results (higher is better).}
\end{minipage}
\end{minipage}
\vspace{-1em}
\setcounter{figure}{7}
\caption{(a) The output of our method running at 12 megapixels, a typical smartphone camera resolution. Despite being trained at a fixed low resolution, our method produces sharp results while faithfully transferring the style from a significantly different scene. See the supplement for full-resolution images. (b) Performance benchmarks on a NVIDIA Tesla V100 GPU with 16 GB of RAM. OOM indicates out of memory. Note that photorealistic postprocessing adds significant overhead to LST performance. Due to GPU loading and startup time, we were unable to get a precise measurement below 5ms. (c) Mean user study scores from 1200 responses. Raters scored the three output images in each sextet on a scale of 1-5 (higher is better).}
\label{fig:hd_runtime_user}
\vspace{-0.2cm}
\end{figure}

%-------------------------------------------------------------------------------

\subsubsection{Visual Comparison.}
We compare our technique against three state-of-the-art photorealistic style transfer algorithms: PhotoWCT~\cite{li2018closed}, LST~\cite{li2019learning}, and WCT\textsuperscript{2}~\cite{yoo2019photorealistic}, using default settings.
Note that for PhotoWCT, we use NVIDIA's latest FastPhotoStyle library.
Comparisons with other algorithms are included in the supplementary material.

Figure~\ref{fig:qualitative} features a small sampling of the test set with some challenging examples.
Owing to its reliance on unpooling and postprocessing, PhotoWCT results contain noticeable artifacts on nearly all scenes. LST mainly focuses on artistic style transfer, and to generate photorealistic results, it uses a compute-intensive spatial propagation network as a postprocessing step to reduce distortion artifacts.
Figure~\ref{fig:qualitative} shows that there are still noticeable distortions in several instances, even after postprocessing.
WCT\textsuperscript{2} performs quite well when content and style are semantically similar, but when the scene content is significantly different from the landscapes on which it was trained, the results appear ``hazy''.
Our method performs well even on these challenging cases.
Thanks to its restricted output space, our method always produce sharp images which degrades gracefully towards the input (e.g., face, leaves) when given inputs outside the training set.
Our primary artifact is a noticeable reduction in contrast along strong edges and is a known limitation of the local affine transform model~\cite{chen2016bilateral}.

%------------------------------------------------------------------------------------
\subsubsection{Robustness.}
Thanks to its restricted transform model, our method is significantly more robust than the baselines when confronted with adversarial inputs, as shown in Figure~\ref{fig:extreme}. Although our model was trained exclusively on landscapes, the restricted transform model allows it to degrade gracefully on portraits which it has never encountered and even a monochromatic ``style''.

%------------------------------------------------------------------------------------

\subsection{Quantitative Results}

\subsubsection{Runtime and Resolution.}
As shown in Figure~\ref{fig:hd_runtime_user}(b), our runtime on a workstation GPU significantly outperforms the baselines and is essentially invariant to resolution at practical resolutions.
This is due to the fact that coefficient prediction, the ``deep'' part of the network runs at a constant low resolution of $256 \times 256$.
In contrast, our full-resolution stream does minimal work and has hardware acceleration for trilinear interpolation.
On a modern smartphone GPU, inference runs comfortably above 30 Hz at full 12 megapixel camera resolution when quantized to 16-bit floating point.
Figure~\ref{fig:hd_4k} shows one such example.
More images and a detailed performance benchmark are included in the supplement.

%-------------------------------------------

\subsubsection{User Study.}
The question of whether an image is a faithful rendition of the style of another is inherently a matter of subjective taste.
As such, we conducted a user study to judge whether our method delivers subjectively better results compared to the baselines.
We recruited 20 users unconnected with the project.
Each user was shown 20 sextets of images consisting of the input content, reference style, and four randomly shuffled outputs (PhotoWCT~\cite{li2018closed}, WCT\textsuperscript{2}~\cite{yoo2019photorealistic}, LST~\cite{li2019learning}, and ours).
For each output, they were asked to rate the following questions on a scale of 1--5:
\vspace{-0.5em}
\begin{itemize}
    \item How noticeable are artifacts (i.e., less photorealistic) in the image?
    \item How similar is the output in style to the reference?
    \item How would you rate the overall quality of the generated image?
\end{itemize}
In total, we collected 1200 responses (400 images $\times$ 3 questions). As the results shown in Figure~\ref{fig:hd_runtime_user}(c) indicate, WCT\textsuperscript{2} achieves similar average scores to our results in terms of photorealism, and both results are significantly better than PhotoWCT. However, in terms of both stylization and overall quality, our technique outperforms all the other related work: PhotoWCT, LST, and WCT\textsuperscript{2}.

%--------------------------------------------------------------------------------------------------------

\subsubsection{Video Stylization.}
\begin{figure}[ht]
  \centering
  \includegraphics[width = 0.95\columnwidth,height=.16\linewidth]{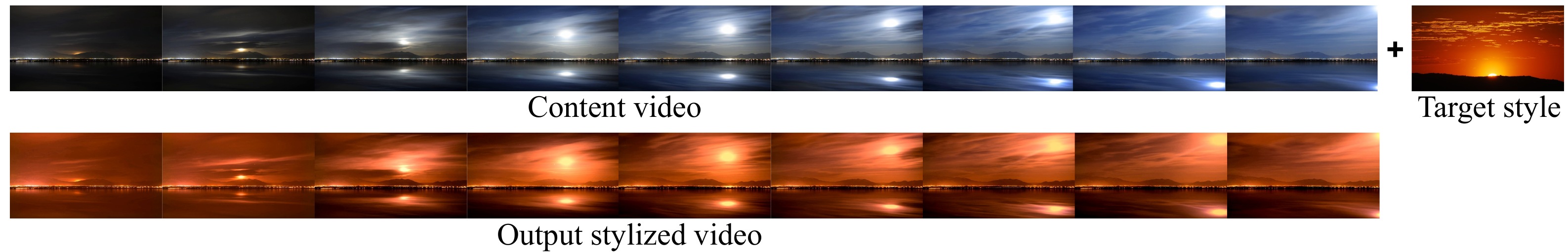}

  \caption{Transferring the style of a still photo to a video sequence. Although the content frames undergo substantial changes in appearance, our method produces a temporally coherent result consistent with the reference style. Please refer to the supplementary material for the full videos.}
  \label{fig:video}

\end{figure}

Although our network is trained exclusively on images, it generalizes well to video content.
Figure~\ref{fig:video} shows an example where we transfer the style of a single photo to a video sequence that varies dramatically in appearance.
The resulting video has a consistent style and is temporally coherent without any additional regularization or data augmentation.
% The full video sequences are in the supplement.

%--------------------------------------------------------------------------------------------------------

\section{Conclusion}
We presented a feed-forward neural network for universal photorealistic style transfer.
The key to our approach is using deep learning to predict affine bilateral grids, which are compact image-to-image transformations that implicitly enforce the photorealism constraint.
We showed that our technique is significantly faster than state of the art, runs in real-time on a smartphone, and degrades gracefully even in extreme cases.
We believe its robustness and fast runtime will lead to practical applications in mobile photography.
As future work, we hope to further improve performance by reducing network size, and investigate how to relax the photorealism constraint to generate a continuum between photorealistic and abstract art.

%-----------------------------------------------------

\begin{figure}[ht]
\centering
\begin{tabular}{c@{\hspace{0.002\linewidth}}c@{\hspace{0.002\linewidth}}c@{\hspace{0.002\linewidth}}c@{\hspace{0.002\linewidth}}c@{\hspace{0.002\linewidth}}c}

\includegraphics[width = .13\linewidth,height=.14\linewidth]{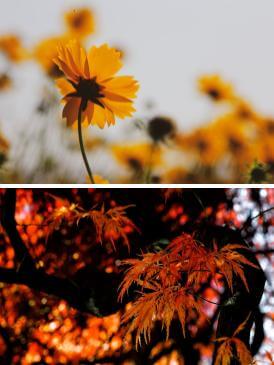} &
   \includegraphics[width = .22\linewidth,height=.14\linewidth]{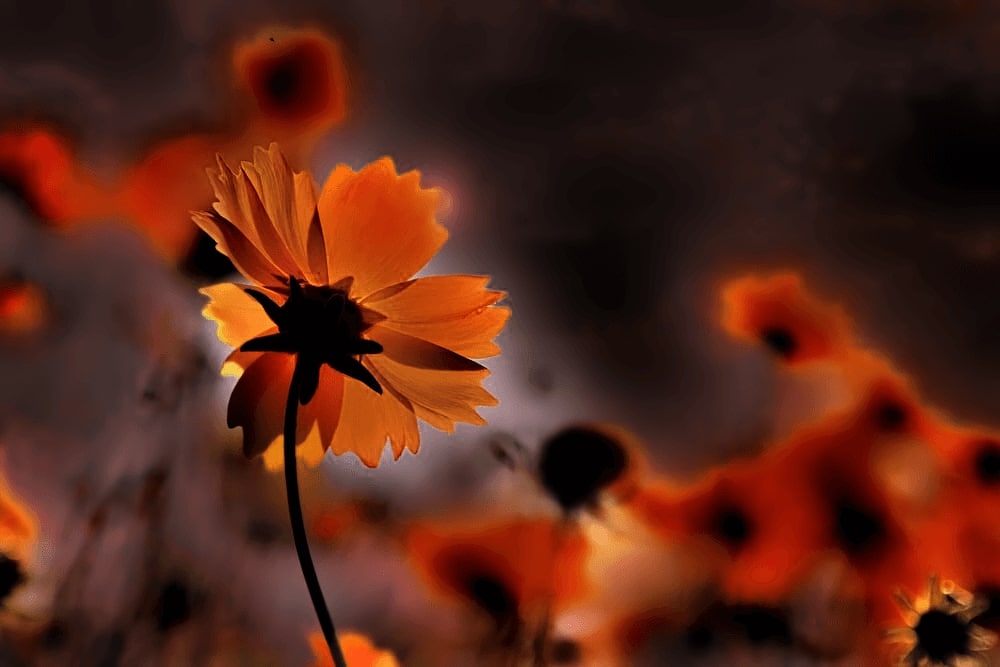} &
  \includegraphics[width = .22\linewidth,height=.14\linewidth]{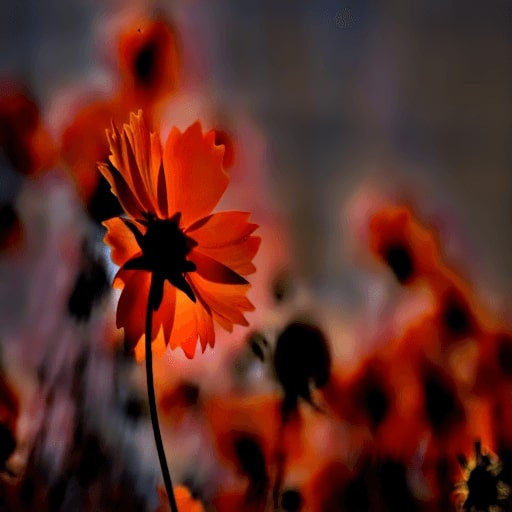} &
   \includegraphics[width = .22\linewidth,height=.14\linewidth]{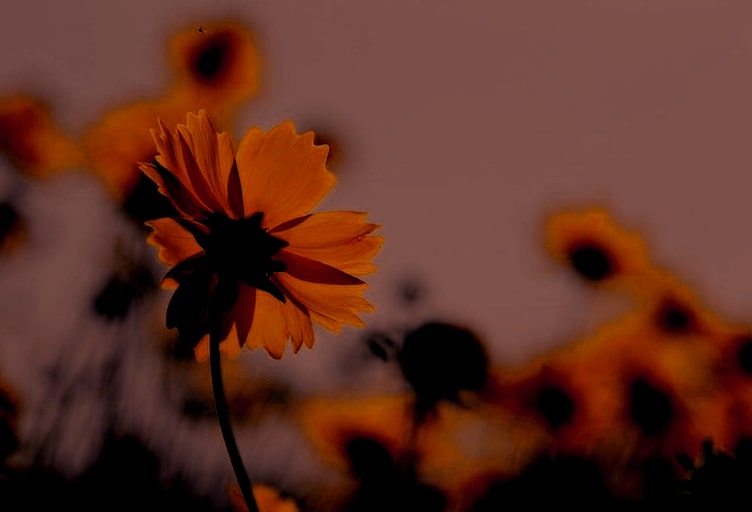} &
   \includegraphics[width = .22\linewidth,height=.14\linewidth]{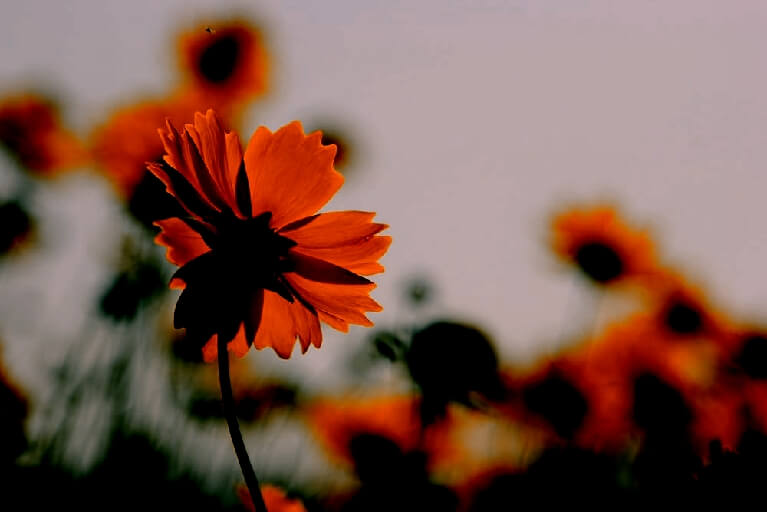} &\\
      \vspace{-0.1em}

    \includegraphics[width = .13\linewidth,height=.14\linewidth]{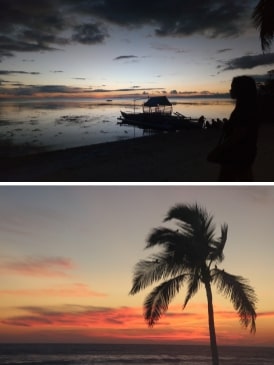} &
   \includegraphics[width = .22\linewidth,height=.14\linewidth]{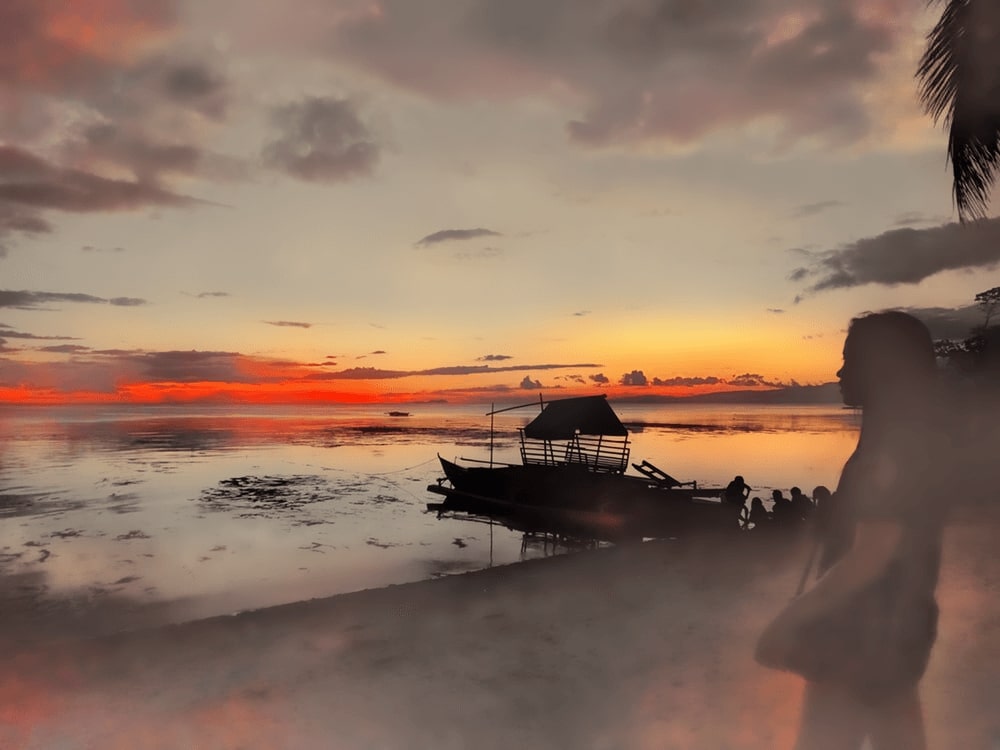} &
  \includegraphics[width = .22\linewidth,height=.14\linewidth]{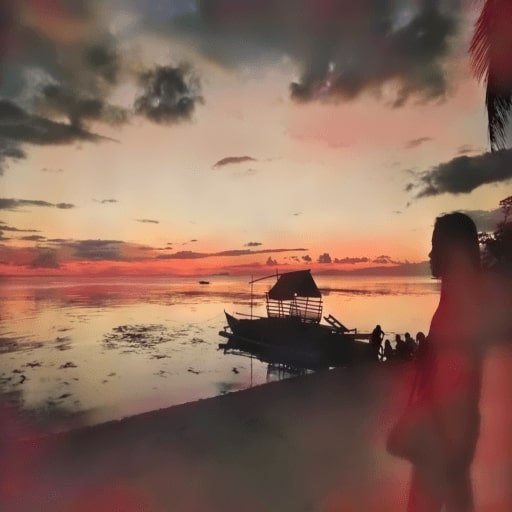} &
   \includegraphics[width = .22\linewidth,height=.14\linewidth]{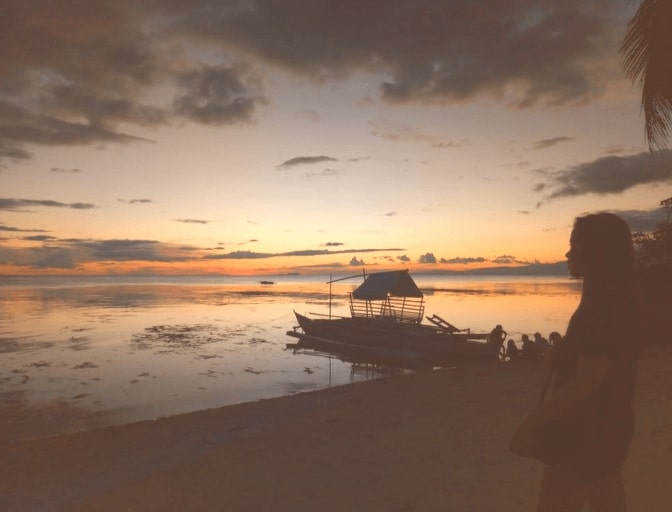} &
   \includegraphics[width = .22\linewidth,height=.14\linewidth]{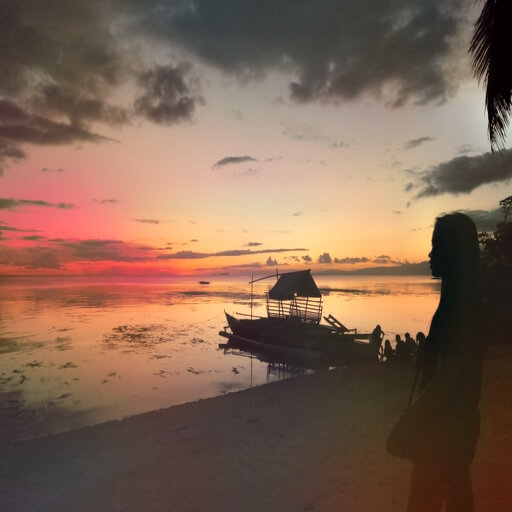} &\\
    \vspace{-0.1em}

\includegraphics[width = .13\linewidth,height=.14\linewidth]{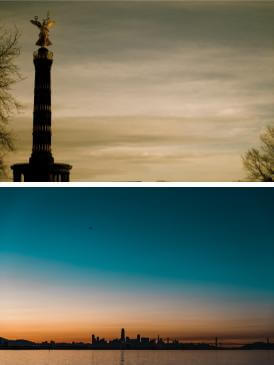} &
   \includegraphics[width = .22\linewidth,height=.14\linewidth]{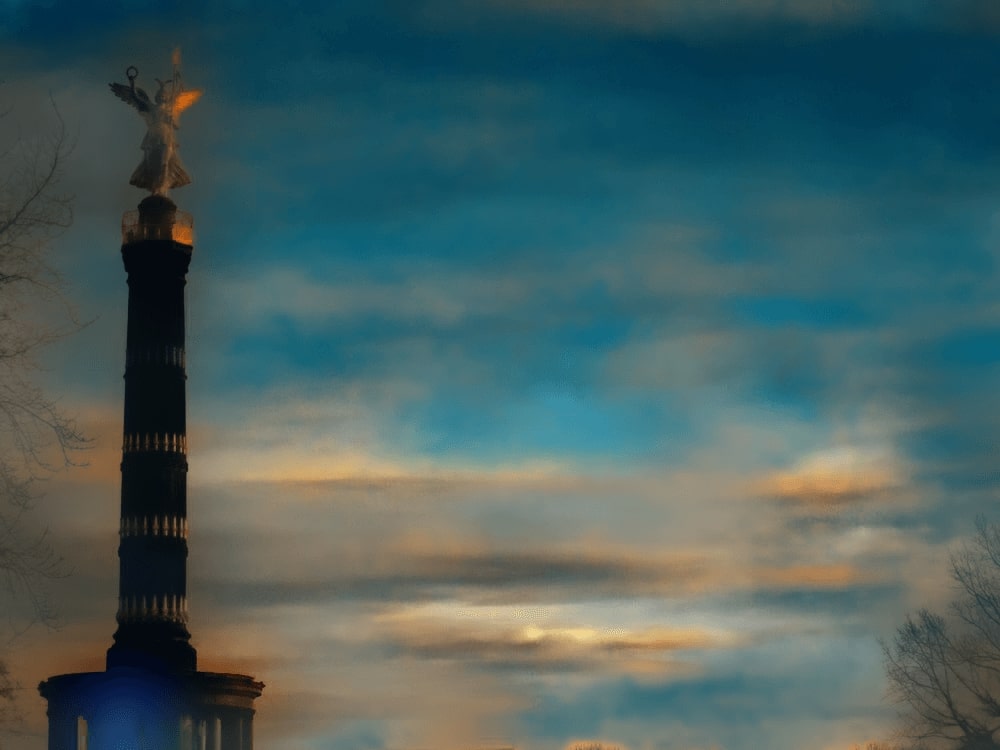} &
  \includegraphics[width = .22\linewidth,height=.14\linewidth]{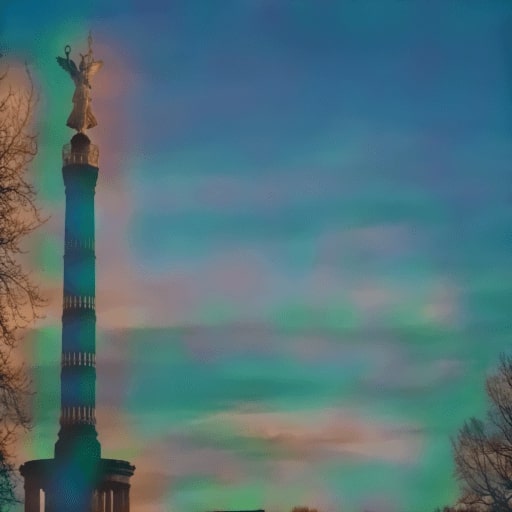} &
   \includegraphics[width = .22\linewidth,height=.14\linewidth]{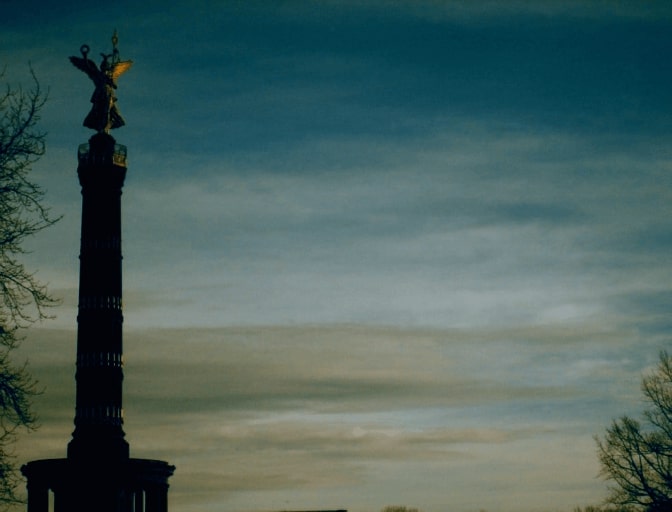} &
   \includegraphics[width = .22\linewidth,height=.14\linewidth]{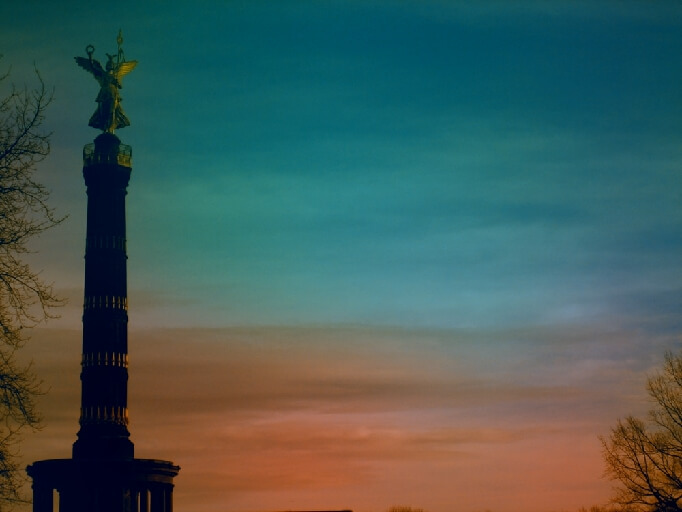} &\\
      \vspace{-0.1em}

   \includegraphics[width = .13\linewidth,height=.14\linewidth]{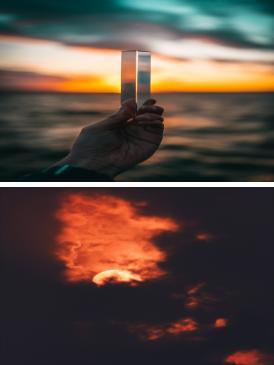} &
   \includegraphics[width = .22\linewidth,height=.14\linewidth]{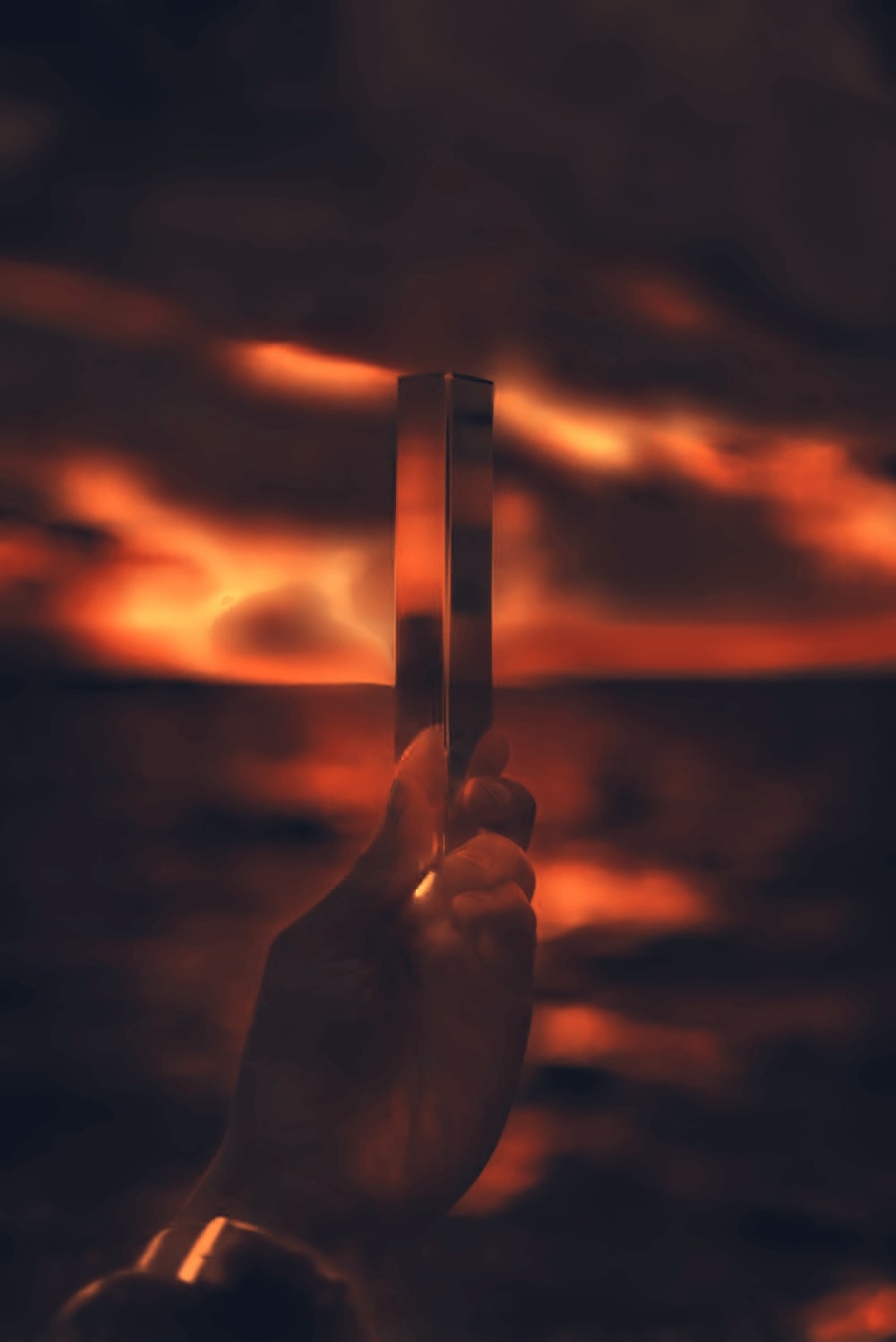} &
  \includegraphics[width = .22\linewidth,height=.14\linewidth]{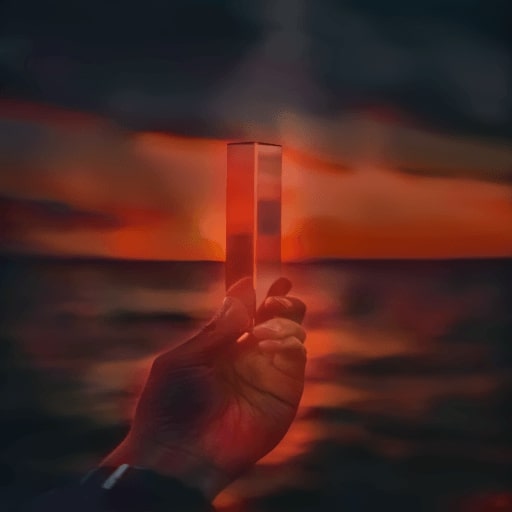} &
   \includegraphics[width = .22\linewidth,height=.14\linewidth]{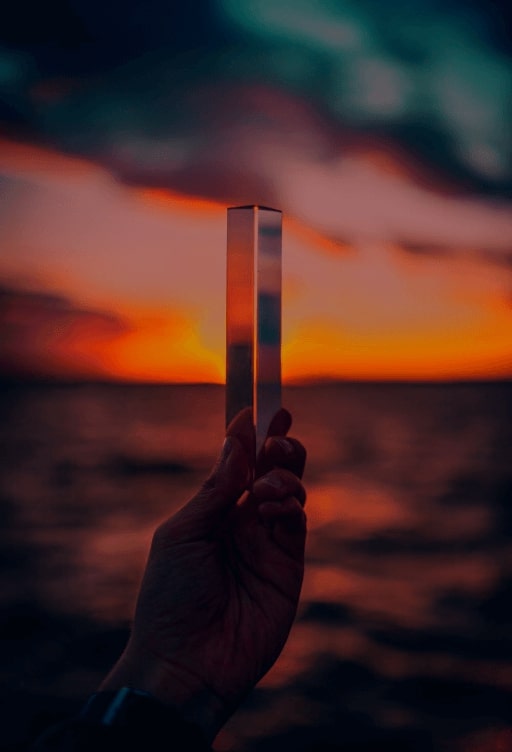} &
   \includegraphics[width = .22\linewidth,height=.14\linewidth]{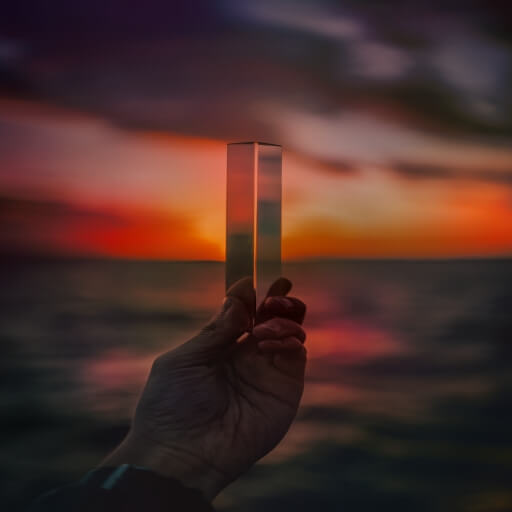} &\\
      \vspace{-0.1em}
   
\includegraphics[width = .13\linewidth,height=.14\linewidth]{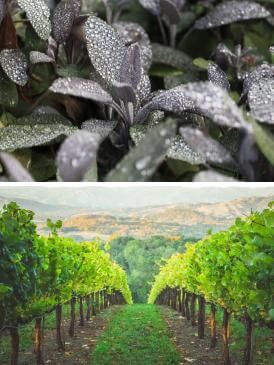} &
   \includegraphics[width = .22\linewidth,height=.14\linewidth]{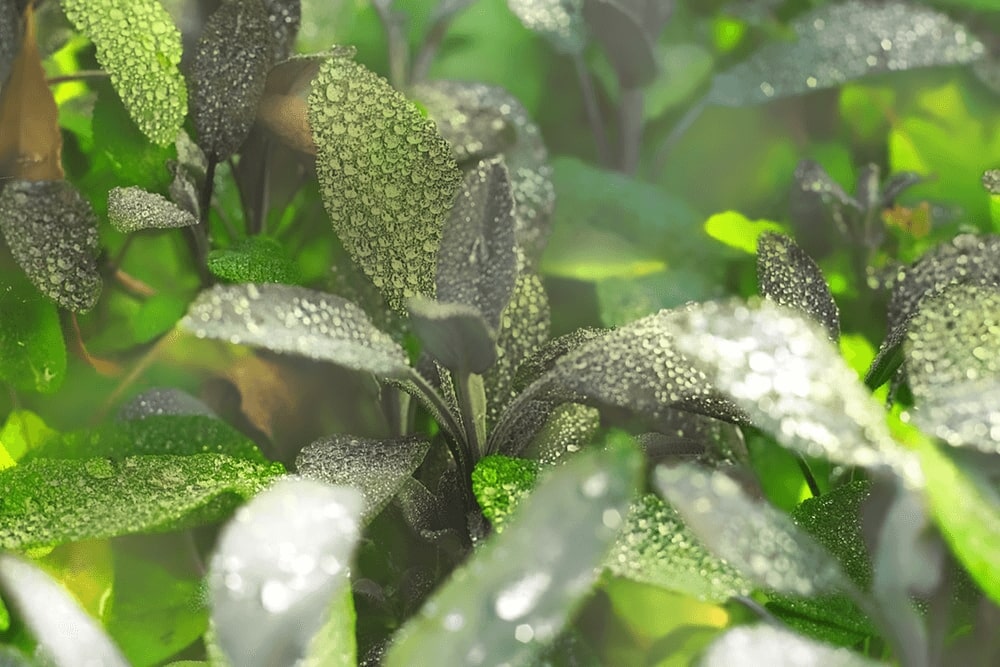} &
  \includegraphics[width = .22\linewidth,height=.14\linewidth]{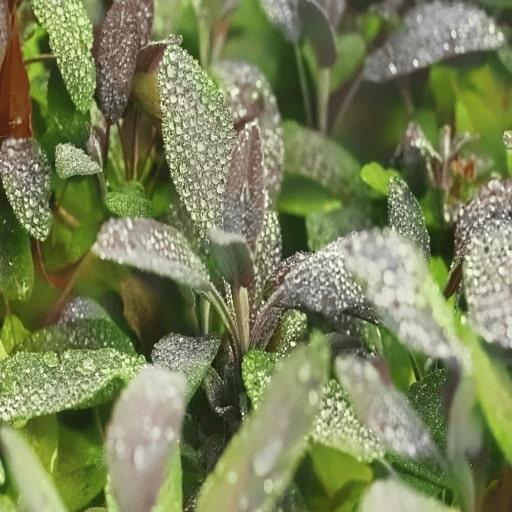} &
   \includegraphics[width = .22\linewidth,height=.14\linewidth]{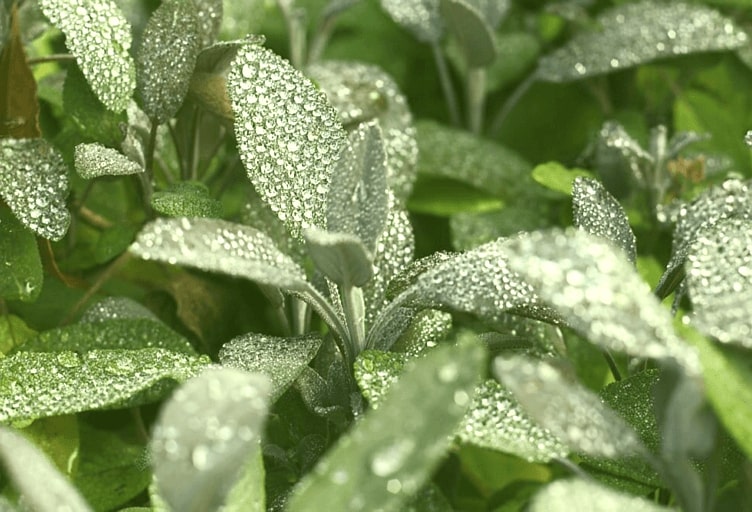} &
   \includegraphics[width = .22\linewidth,height=.14\linewidth]{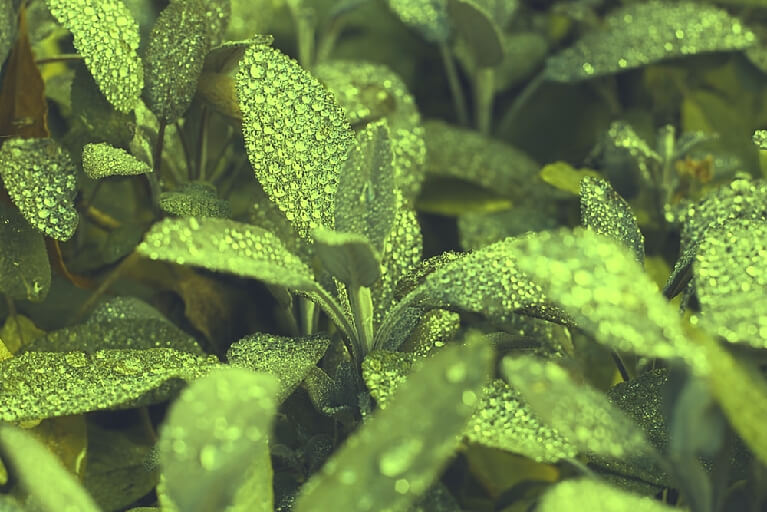} &\\
      \vspace{-0.1em}

    \includegraphics[width = .13\linewidth,height=.14\linewidth]{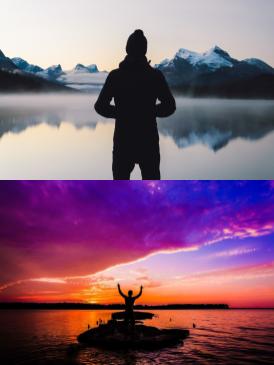} &
   \includegraphics[width = .22\linewidth,height=.14\linewidth]{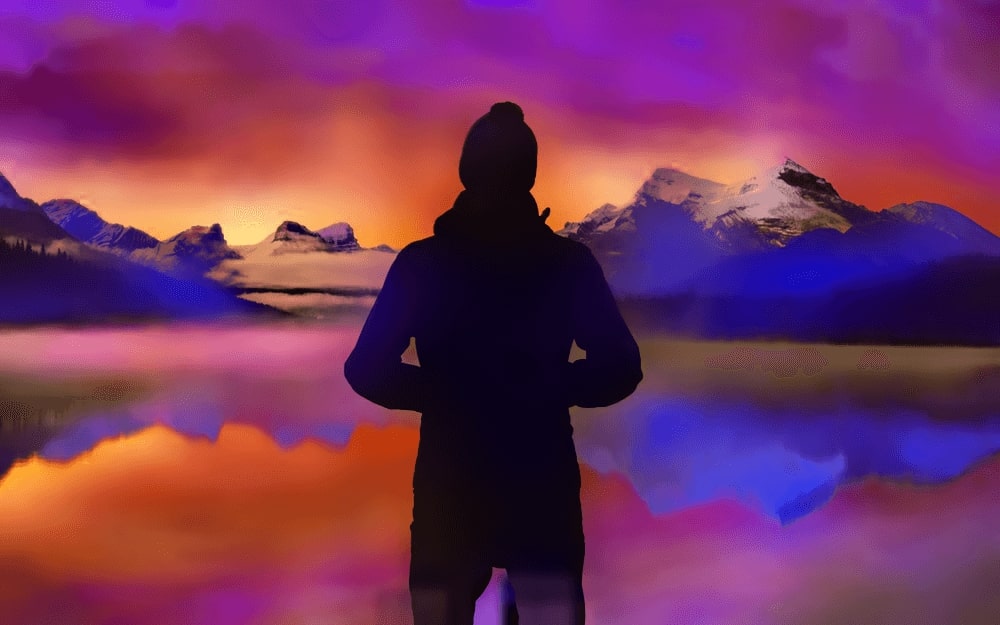} &
  \includegraphics[width = .22\linewidth,height=.14\linewidth]{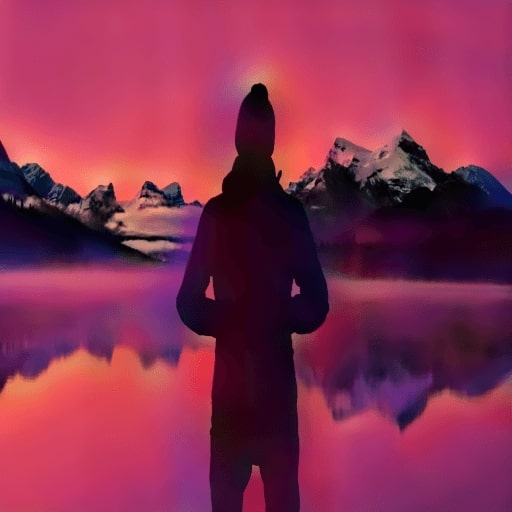} &
   \includegraphics[width = .22\linewidth,height=.14\linewidth]{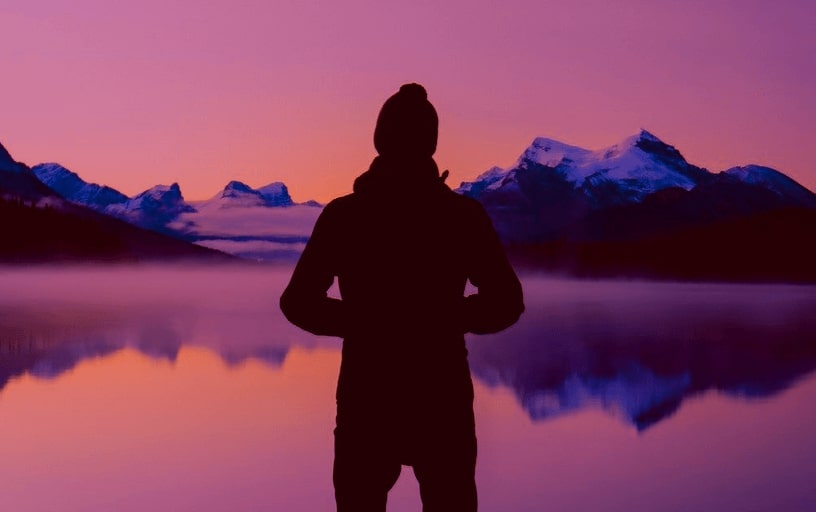} &
   \includegraphics[width = .22\linewidth,height=.14\linewidth]{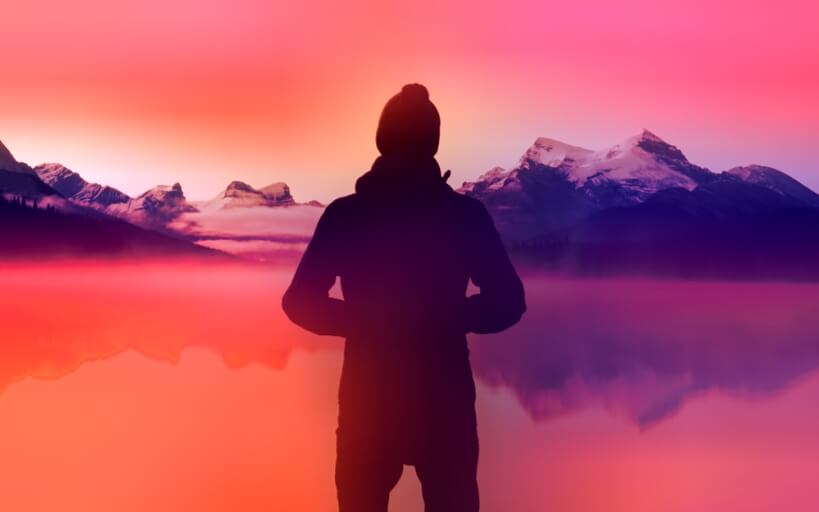} &\\
      \vspace{-0.1em}
      
    \includegraphics[width = .13\linewidth,height=.14\linewidth]{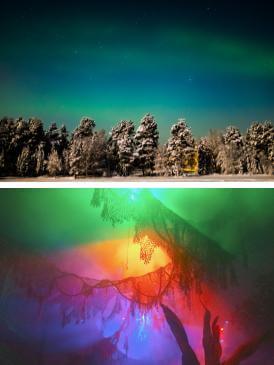} &
   \includegraphics[width = .22\linewidth,height=.14\linewidth]{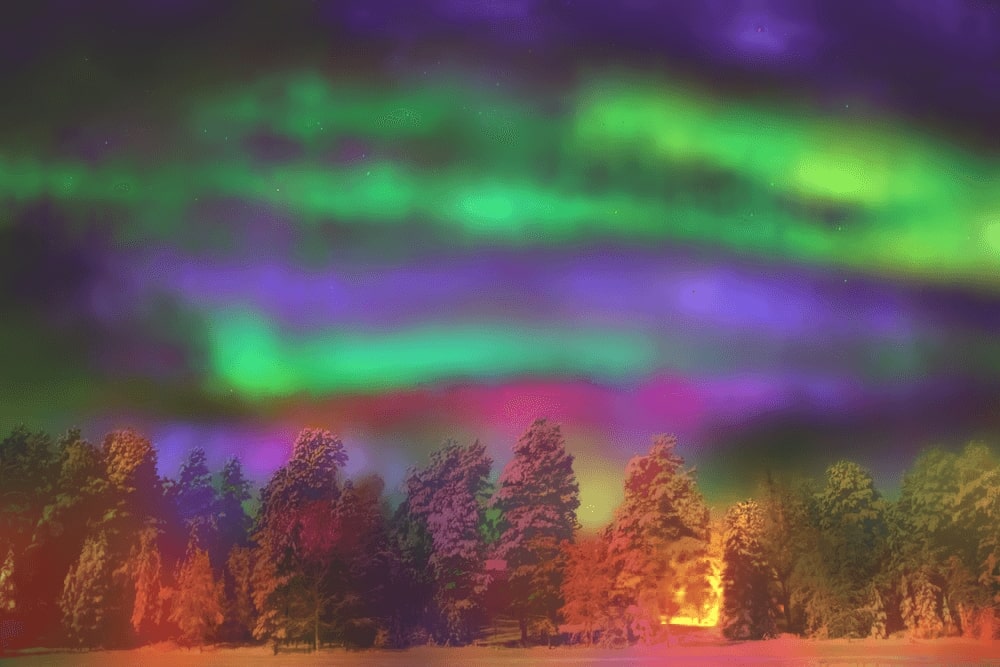} &
  \includegraphics[width = .22\linewidth,height=.14\linewidth]{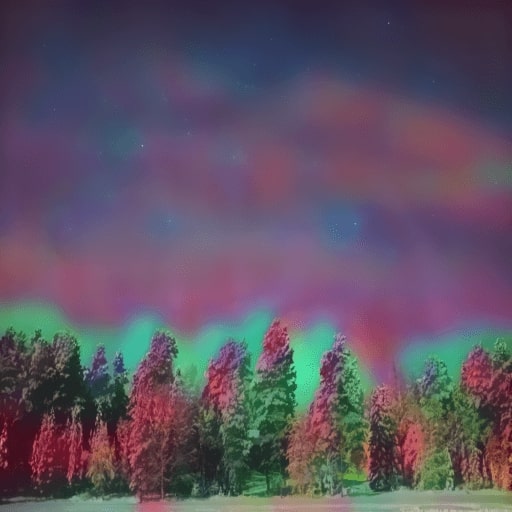} &
   \includegraphics[width = .22\linewidth,height=.14\linewidth]{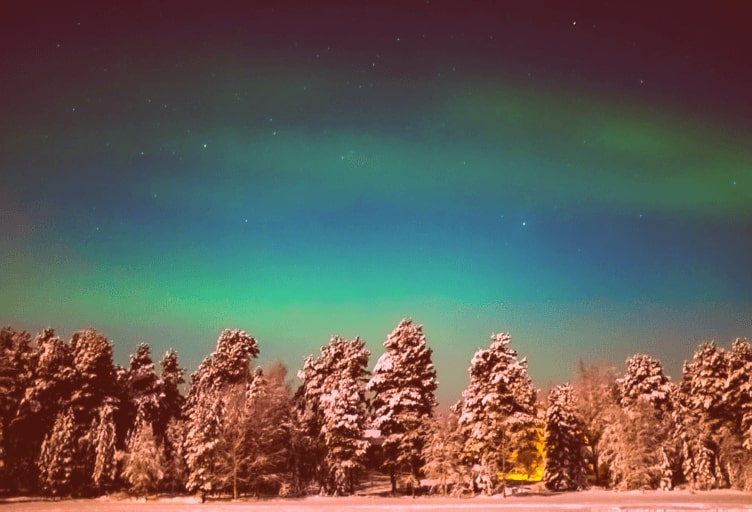} &
   \includegraphics[width = .22\linewidth,height=.14\linewidth]{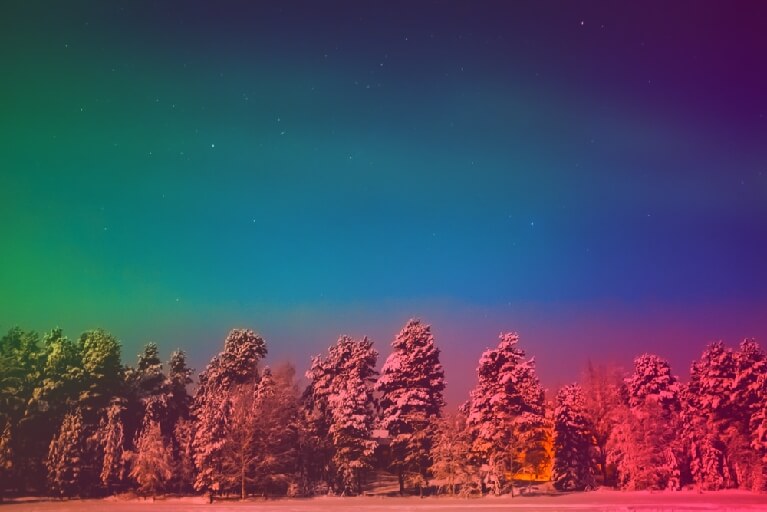} &\\
     \vspace{-0.1em}
   
\includegraphics[width = .13\linewidth,height=.14\linewidth]{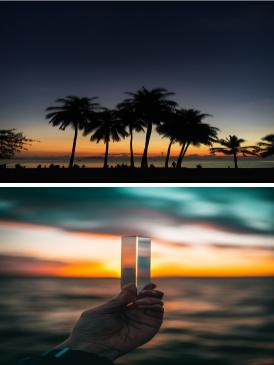} &
   \includegraphics[width = .22\linewidth,height=.14\linewidth]{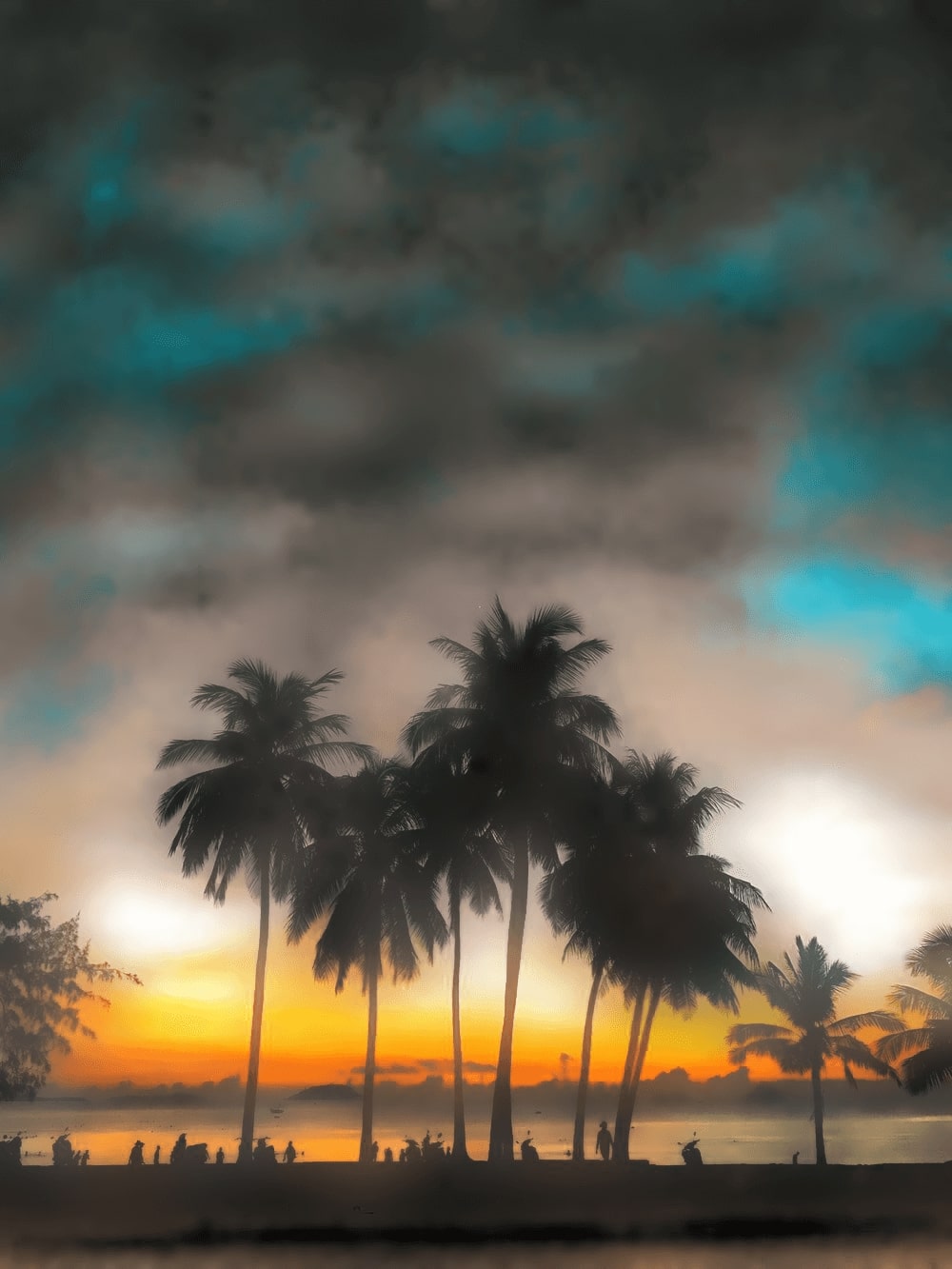} &
  \includegraphics[width = .22\linewidth,height=.14\linewidth]{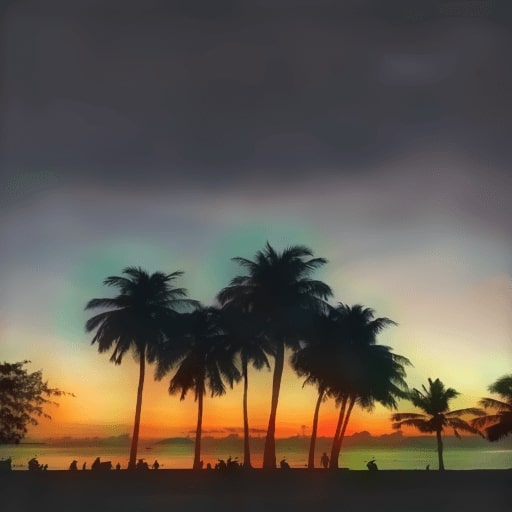} &
   \includegraphics[width = .22\linewidth,height=.14\linewidth]{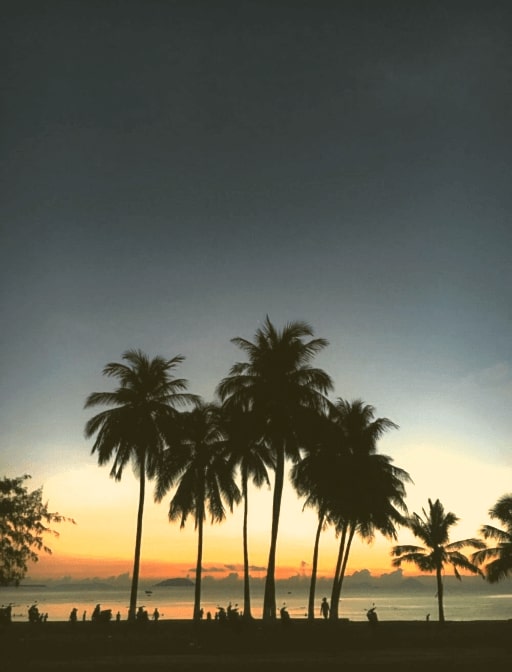} &
   \includegraphics[width = .22\linewidth,height=.14\linewidth]{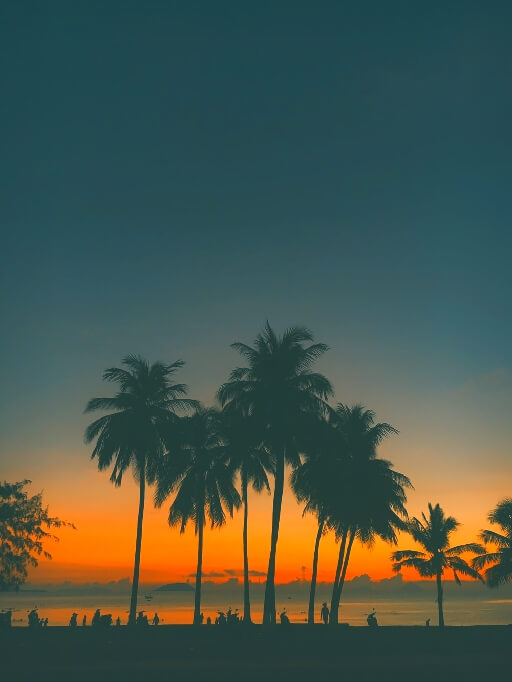} &\\
    \vspace{-0.1em}

    \includegraphics[width = .13\linewidth,height=.14\linewidth]{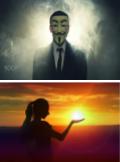} &
   \includegraphics[width = .22\linewidth,height=.14\linewidth]{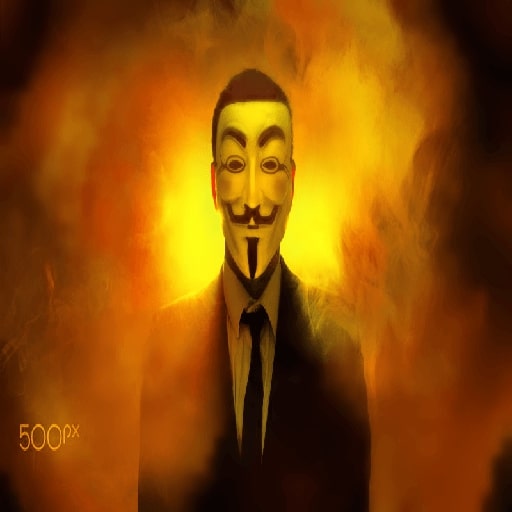} &
  \includegraphics[width = .22\linewidth,height=.14\linewidth]{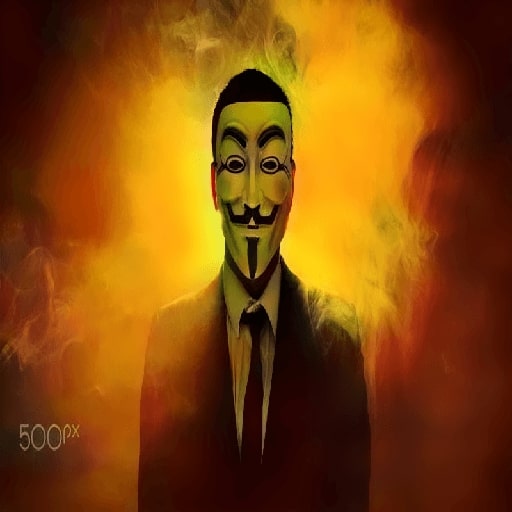} &
   \includegraphics[width = .22\linewidth,height=.14\linewidth]{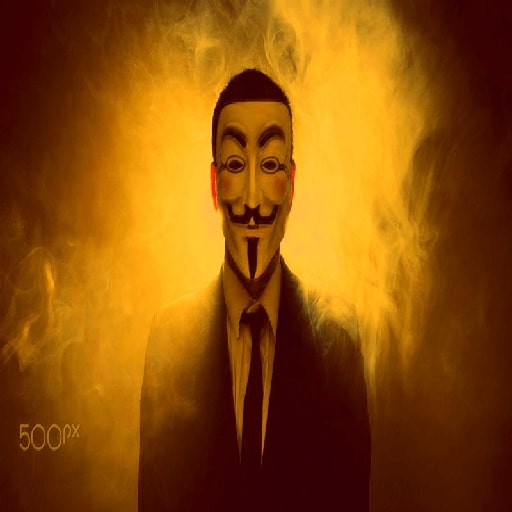} &
   \includegraphics[width = .22\linewidth,height=.14\linewidth]{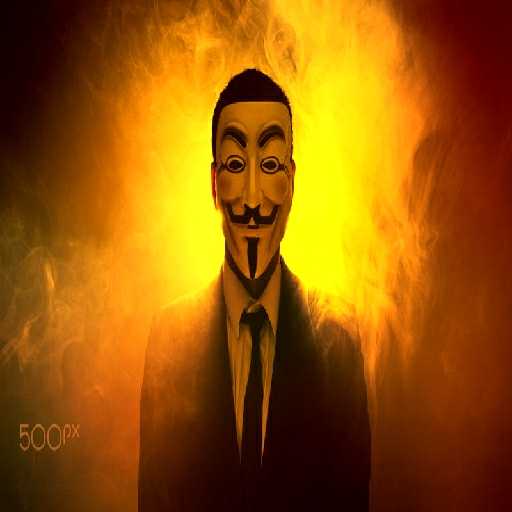} &\\
     \vspace{-0.1em}

        \includegraphics[width = .13\linewidth,height=.14\linewidth]{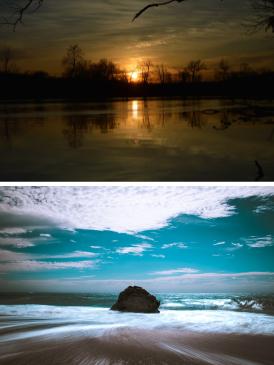} &
   \includegraphics[width = .22\linewidth,height=.14\linewidth]{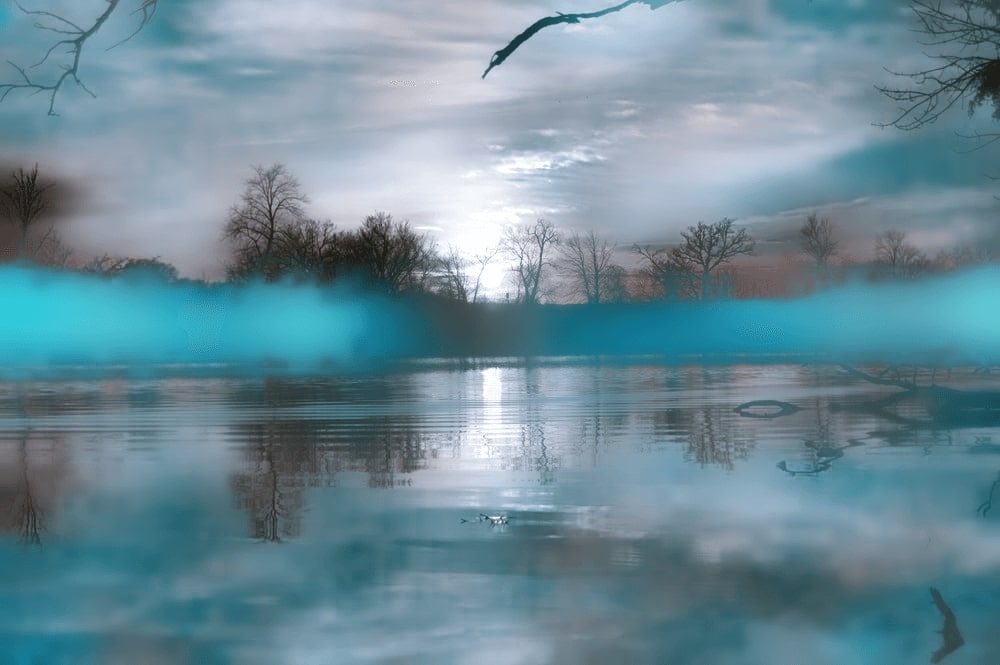} &
  \includegraphics[width = .22\linewidth,height=.14\linewidth]{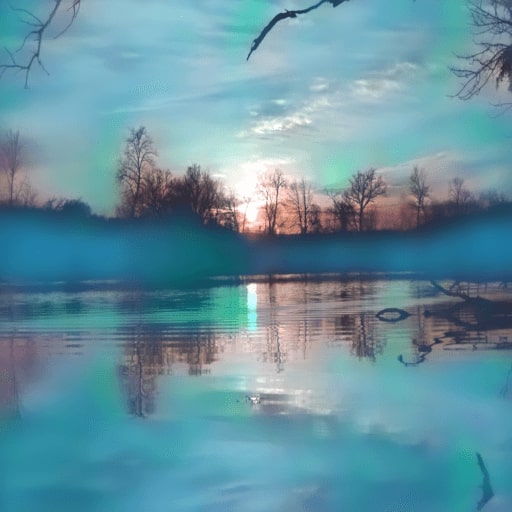} &
   \includegraphics[width = .22\linewidth,height=.14\linewidth]{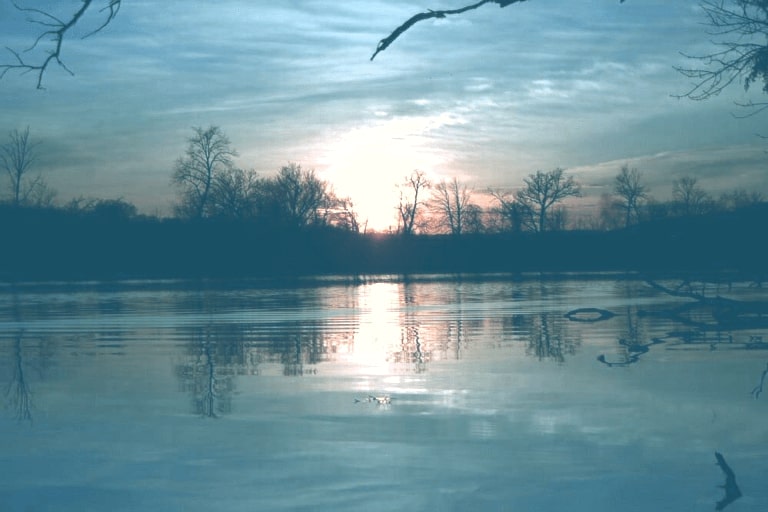} &
   \includegraphics[width = .22\linewidth,height=.14\linewidth]{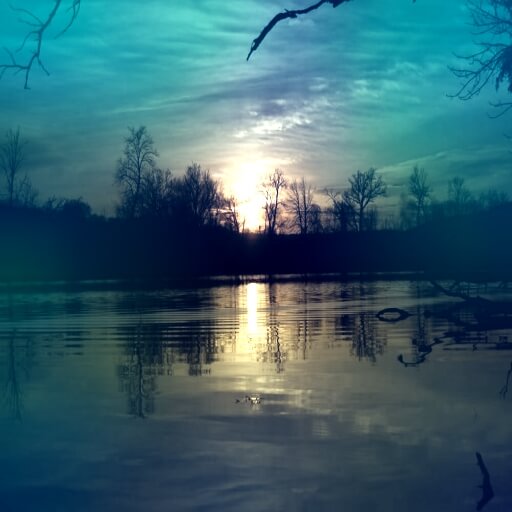} &\\

   \vspace{-1em}
  Inputs & {\small PhotoWCT~\cite{li2018closed}} & LST~\cite{li2019learning} & WCT\textsuperscript{2}~\cite{yoo2019photorealistic} & Ours \\
\end{tabular}
\vspace{-0.1em}
\caption{\textbf{Qualitative comparison} of our method against three state of the art baselines on some challenging examples.}
\label{fig:qualitative}
\end{figure}

%------------------------------------------------------------------------

\clearpage

\bibliographystyle{splncs04}
\bibliography{pst}

\clearpage

\end{document}